\documentclass[10pt,twocolumn,letterpaper]{article}

\usepackage{iccv}
\usepackage{times}
\usepackage{epsfig}
\usepackage{graphicx}
\usepackage{amsmath}
\usepackage{amssymb}

\usepackage{booktabs}
\usepackage{makecell}
\usepackage{float}
\usepackage{color}
\usepackage{dsfont}
\usepackage{mathrsfs}
\usepackage{bbm}
\usepackage{subfig}

\usepackage[pagebackref=true,breaklinks=true,letterpaper=true,colorlinks,bookmarks=false]{hyperref}

\iccvfinalcopy 

\def\iccvPaperID{1593} 

\begin{document}

\title{Parallax-Tolerant Unsupervised Deep Image Stitching}

\author{Lang Nie$^{1,2}$, Chunyu Lin$^{1,2}$\thanks{Corresponding author}, Kang Liao$^{1,2}$, Shuaicheng Liu$^{3}$, Yao Zhao$^{1,2}$\\
$^{1}$Institute of Information Science, Beijing Jiaotong University, Beijing, China\\
$^{2}$Beijing Key Laboratory of Advanced Information Science and Network, Beijing, China\\
$^{3}$University of Electronic Science and Technology of China, Chengdu, China\\
}


\twocolumn[{%
\maketitle
\begin{figure}[H]
\hsize=\textwidth 
\centering
\vspace{-0.9cm}
\includegraphics[width=.92\textwidth, height=.30\textheight]{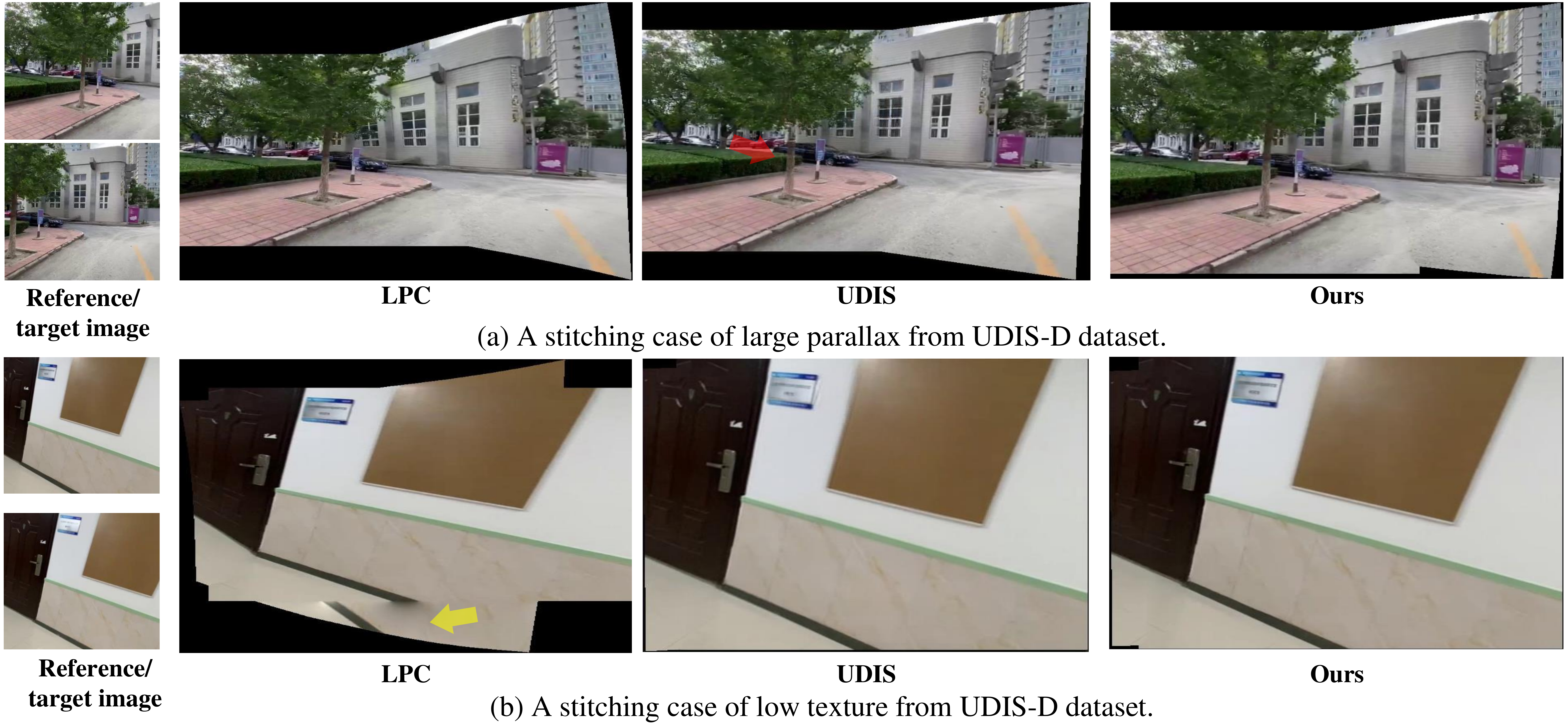}
\vspace{-0.2cm}
\caption{Limitations of existing methods. (a) UDIS \cite{nie2021unsupervised} (learning method) deals with large parallax by blurring parallax regions (highlighted in red). (b) LPC \cite{jia2021leveraging} (traditional method) fails in low-texture scenes without sufficient geometric features. Instead, our solution is free from these limitations, achieving promising results in both of the challenging circumstances.}
\label{fig1}

\end{figure}
}]

\begin{abstract}
\vspace{-0.3cm}
  Traditional image stitching approaches tend to leverage increasingly complex geometric features (\textit{e.g.}, point, line, edge, etc.) for better performance.
  However, these hand-crafted features are only suitable for specific natural scenes with adequate geometric structures.
  In contrast, deep stitching schemes overcome adverse conditions by adaptively learning robust semantic features, but they cannot handle large-parallax cases.


  To solve these issues, we propose a parallax-tolerant unsupervised deep image stitching technique (UDIS++). First, we propose a robust and flexible warp to model the image registration from global homography to local thin-plate spline motion. It provides accurate alignment for overlapping regions and shape preservation for non-overlapping regions by joint optimization concerning alignment and distortion.
  Subsequently, to improve the generalization capability, we design a simple but effective iterative strategy to enhance the warp adaption in cross-dataset and cross-resolution applications.
  Finally, to further eliminate the parallax artifacts, we propose to composite the stitched image seamlessly by unsupervised learning for seam-driven composition masks.
  Compared with existing methods, our solution is parallax-tolerant and free from laborious designs of complicated geometric features for specific scenes.
  Extensive experiments show our superiority over the SoTA methods, both quantitatively and qualitatively. The code is available at \url{https://github.com/nie-lang/UDIS2}.
  \end{abstract}

\vspace{-0.7cm}

\section{Introduction}
\label{sec:introduction}
\vspace{-0.1cm}

Image stitching is a practical technology that aims to construct a scene with a wide field-of-view (FoV) from different images with limited FoV. It is useful in a wide range of fields, such as autonomous driving, medical imaging, surveillance videos, virtual reality, etc.

Over the past decades, traditional stitching approaches tend to adopt increasingly complicated geometric features to achieve better content alignment and shape preservation. In the beginning, SIFT \cite{lowe2004distinctive} is widely used in various image stitching algorithms \cite{brown2007automatic, gao2011constructing, zaragoza2013projective, chang2014shape, lin2015adaptive, lee2020warping} to extract discriminative key points and calculate adaptive warps. Then, the line segment is proved to be another unique feature to achieve better stitching quality and preserve linear structures \cite{li2015dual, xiang2018image, liao2019single, jia2021leveraging}. Recently, the large-scale edge is also introduced in \cite{du2022geometric} to preserve the contour structures. Besides, there is a great variety of other geometric features that are leveraged to improve the stitching quality, such as depth maps \cite{liao2022natural}, semantic planar regions \cite{li2021image}, etc.

Having calculated the warps, seam cutting is usually used to remove parallax artifacts. To explore an invisible seam, various energy functions are designed using colors \cite{kwatra2003graphcut}, edges \cite{lin2016seagull, dai2021edge}, salient maps \cite{li2018perception}, depth \cite{chen2022optimized}, etc.

From the broad usage of geometric features, a clear developing trend has been discovered: increasingly sophisticated features are leveraged.
We ask: are these complex designs practical in real applications? We attempt to answer this question from two perspectives. 1) These elaborate algorithms with complicated geometric features poorly adapt to scenes without sufficient geometric structures, such as medical images, industrial images, and other natural images with low texture (Fig.\ref{fig1}{\color{red}b}), low light or low resolution. 2) When there exist abundant geometric structures, the running speed is intolerant (please refer to Table \ref{tab2},\ref{tab3} for detail). Such a trend seems to violate the ``practical" original intent.

Recently, deep stitching technologies using convolutional neural networks (CNNs) have aroused widespread attention in the community. They abandon geometric features and head for high-level semantic features that can be adaptively learned in a data-driven pattern in a supervised \cite{lai2019video, nie2020view, nie2022learning, song2021end,  kweon2021pixel}, weakly-supervised \cite{song2022weakly}, or unsupervised \cite{nie2021unsupervised} manner. Although they are robust to various natural or unnatural conditions, they cannot handle large parallax and demonstrate unsatisfactory generalization in cross-dataset and cross-resolution conditions. A large-parallax case is shown in Fig.\ref{fig1}{\color{red}a}, where the tree is in the middle of the car in the reference image while it is on the left in the target image. To deal with parallax, UDIS \cite{nie2021unsupervised} reconstructs stitched images from feature to pixel. However, the parallax is so large that undesired blurs are produced as a side effect.

In this paper, we propose a parallax-tolerant unsupervised deep image stitching technique, addressing the robustness issue in traditional stitching and the large-parallax issue in deep stitching simultaneously.
Actually, the proposed deep learning-based solution is naturally robust to various scenes due to effective semantic feature extraction. Then, it overcomes the large parallax via two stages: warp and composition.
In the first stage, we propose a robust and flexible warp to model the image registration. Particularly, we simultaneously parameterize homography transformation and thin-plate spline (TPS) transformation as unified representations in a compact framework. The former offers a global linear transformation, while the latter produces local nonlinear deformation, allowing our warp to align images with parallax. Besides, this warp contributes to both content alignment and shape preservation simultaneously via combined optimization of alignment and distortion.
In the second stage, the existing reconstruction-based method \cite{nie2021unsupervised} treats artifact elimination as a reconstruction process from feature to pixel, leading to inevitable blurs around the parallax regions. To overcome this drawback, we cooperate the motivation of seam-cutting into deep composition and implicitly find a ``seam" through unsupervised learning for seam-driven composition masks. 
To this end, we design boundary and smoothness constraints to restrict the endpoints and route of a ``seam", compositing the stitched image seamlessly.
In addition to the two stages, we design a simple iterative strategy to enhance the generalization, rapidly improving the registration performance of our warp in different datasets and resolutions.

Furthermore, we conduct extensive experiments about the warp and composition, demonstrating our superiority to other SoTA solutions.
The contributions center around:
\vspace{-0.2cm}
\begin{itemize}
    \item We propose a robust and flexible warp by parameterizing the homography and thin-plate spline into unified representations, realizing unsupervised content alignment and shape preservation in various scenes.
    \vspace{-0.2cm}
    \item A new composition approach is proposed to generate seamless stitched images via unsupervised learning for composition masks. Compared with the reconstruction \cite{nie2021unsupervised}, our composition eliminates parallax artifacts without introducing undesirable blurs.
    \vspace{-0.2cm}
    \item We design a simple iterative strategy to enhance warp adaption in different datasets and resolutions.
 \end{itemize}

\section{Related Work}
\label{sec:related_work}
\subsection{Traditional Image Stitching}
\textbf{Adaptive warp.}
 AutoStitch \cite{brown2007automatic} leveraged SIFT \cite{lowe2004distinctive} to extract discriminative keypoints to construct a global homography transformation. After that, SIFT becomes an indispensable feature to calculate various flexible warps, such as DHW \cite{gao2011constructing}, SVA \cite{lin2011smoothly} APAP \cite{zaragoza2013projective}, ELA \cite{li2017parallax}, TFA \cite{li2019local} for better alignment, SPHP \cite{chang2014shape}, AANAP \cite{lin2015adaptive}, GSP \cite{chen2016natural} for better shape preservation.
Then, DFW \cite{gao2011constructing} adopted line segments extracted by LSD \cite{von2008lsd} with keypoints together to enrich structural information in artificial environments. Furthermore, line-guided mesh deformation \cite{xiang2018image} is designed by optimizing an energy function of various line-preserving terms \cite{liao2019single, jia2021leveraging}. To preserve the nonlinear structures, the edge features are used in GES-GSP \cite{du2022geometric} to achieve a smooth transition between local alignment and structural preservation. In addition to these basic geometric features (point, line, and edge), the depth maps and semantic planes are also used to assist the feature matching using extra depth consistency \cite{liao2022natural} and planar consensus \cite{li2021image}.

\textbf{Seam cutting.}
The seam cutting is usually used as a post-processing operation to composite stitched images, which introduces an optimization problem of label assignment along the seam. To obtain a plausible stitched result, an extensive range of energy terms are defined by penalizing photometric differences, such as the Euclidean-metric color difference \cite{kwatra2003graphcut}, gradient difference \cite{agarwala2004interactive, dai2021edge},  motion- and exposure-aware difference \cite{eden2006seamless}, salient difference \cite{li2018perception}, etc. Then these energy functions are minimized via graph-cut optimization \cite{kwatra2003graphcut}. Besides that, seam cutting is also applied in image alignment to find the best alignment warp with minimal seam-based cost \cite{gao2013seam, zhang2014parallax, lin2016seagull, li2022automatic}.

These complex geometric features are beneficial in natural scenes with adequate geometric structures. However, there are two drawbacks: 1) Without sufficient geometric structures, the strict feature requirements yield inferior stitching quality, even failure. 2) With excessive geometric structures, the computational cost leaps dramatically.

\vspace{-0.1cm}
\subsection{Deep Image Stitching}
\vspace{-0.15cm}
In contrast, deep stitching schemes are free from endless designs of geometric features. They learn to capture high-level semantic features from extensive data automatically in a supervised \cite{lai2019video, nie2020view, nie2022learning, song2021end,  kweon2021pixel}, weakly-supervised \cite{song2022weakly}, or unsupervised \cite{nie2021unsupervised} fashion, making them robust to various challenging scenes. Among them, the unsupervised one \cite{nie2021unsupervised} is more popular due to the unavailability of real stitched labels. However, it cannot handle large parallax due to the limitation of the homography-based alignment model. The subsequent reconstruction would bring undesirable blurs around parallax regions.

\begin{figure*}[!t]
  \centering
  \includegraphics[width=0.95\textwidth]{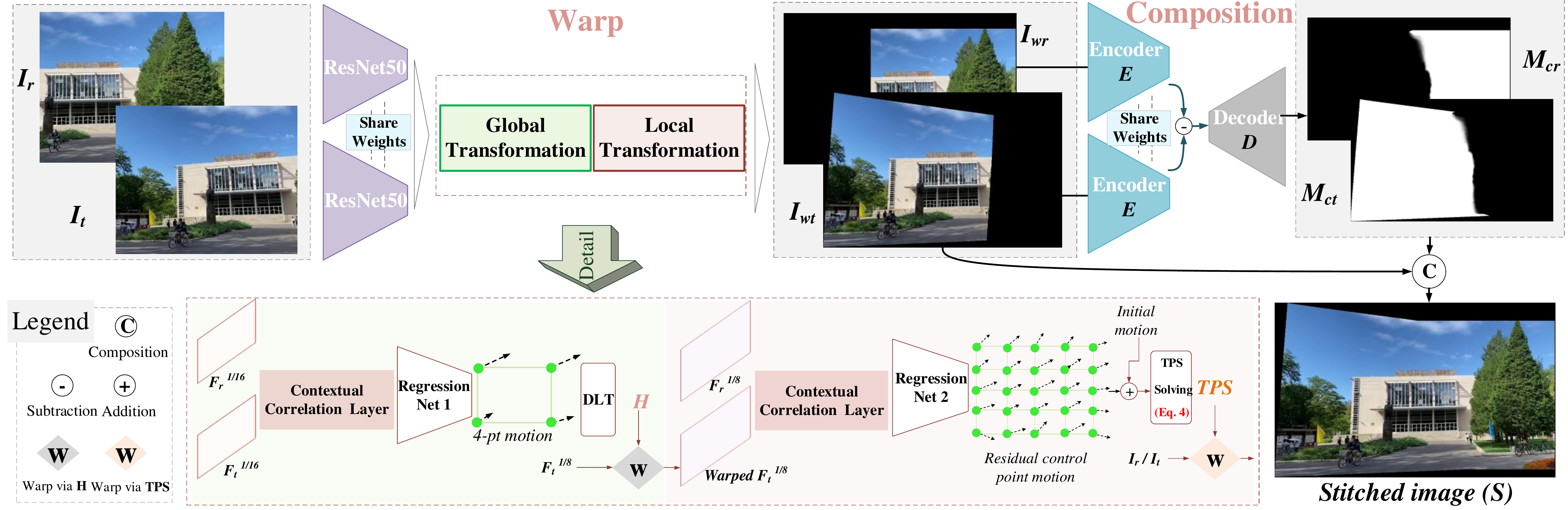}
  \vspace{-3pt}
  \caption{An overview of the proposed parallax-tolerant unsupervised stitching network. Our framework consists of two stages: warp and composition. The first stage predicts a robust and flexible warp to align images with shape preservation. The second stage composites the seamless stitched image by generating composition masks corresponding to warped images.}
  \label{fig:network}
  \vspace{-0.3cm}
\end{figure*}

\vspace{-3pt}

\section{Methodology}
\label{sec:methodology}
The overview of our method is shown in Fig.\ref{fig:network}, where the proposed framework is composed of two stages: warp and composition. In the first stage, our method takes a reference image ($I_r$) and a target image ($I_t$) with overlapping regions as input, and regresses a robust and flexible warp. Then the warped images ($I_{wr}, I_{wt}$) are input to the second stage to predict composition masks ($M_{cr}, M_{ct}$). The stitched image ($S$) can be seamlessly composited as follows:
\begin{eqnarray}
  S = M_{cr} \times I_{wr}  + M_{ct} \times I_{wt}.
\end{eqnarray}

\subsection{Unsupervised Warp Construction}
\label{sec:31}

\subsubsection{Warp Parameterization}
\label{sec:311}
\vspace{-0.1cm}

The homography transformation is an invertible mapping from one image to another with 8 degrees of freedom: each two for translation, rotation, scale, and lines at infinity. To guarantee the non-singularity \cite{nguyen2018unsupervised} in a regression network, it is commonly parameterized as the motions of four vertices \cite{detone2016deep}, which is solved as a $3\times 3$ matrix using DLT \cite{hartley2003multiple}. 

However, if a non-planar scene is captured by cameras with different shooting centers, the homography fails to achieve accurate alignment. To solve it, the mesh-based multi-homography scheme \cite{zaragoza2013projective} is usually used in traditional stitching algorithms. But it cannot be efficiently parallel accelerated, which means it fails to be used in a deep learning framework \cite{9605632, nie2022deep}. \textit{Please refer to Section 2.3 of the supplementary material for specific analysis.} To overcome this issue, we propose to leverage TPS transformation \cite{bookstein1989principal, jaderberg2015spatial} to achieve efficient local deformation.

TPS transformation is a nonlinear, flexible transformation that is usually used to approximate the deformation of non-rigid objects using a thin plate. It is determined by two sets of control points, with a one-to-one correspondence between a flat image and a warped image. Denote $N$ control points on a flat image as $P=[p_1,...,p_N]^T$ and corresponding points on the warped image as $P'=[p'_1,...,p'_N]^T$ ($p_i,p'_i\in \mathbb{R}^{2\times 1}$). By minimizing an energy function consisting of a data term and a distortion term \cite{kent1994link} (\textit{refer to Section 2.1 of the supplementary material for more details}), the TPS transformation can be parameterized as Eq.\ref{eq2}:
\begin{equation}
  p'=\mathscr{T}(p)=C+Mp+\sum_{i=1}^{N}w_iO(\parallel p-p_i\parallel_2 ),
  \label{eq2}
\end{equation}
where $p$ is an arbitrary point on the flat image and $p'$ is the corresponding point on the warped image. $C\in \mathbb{R}^{2\times 1}$, $M\in \mathbb{R}^{2\times 2}$, and $w_i\in \mathbb{R}^{2\times 1}$ are the transformation parameters. $O(r)=r^2logr^2$ is a radial basis function that indicates the impact of each control point on $p$. To solve these parameters, we formulate $N$ data constraints using $N$ pairs of control points according to Eq.\ref{eq2}, and impose extra dimensional constraints \cite{kent1994link} as described in Eq.\ref{eq3}:
\begin{equation}
  \sum_{i=1}^{N}w_i=0 \hspace{0.2cm} and \hspace{0.2cm} \sum_{i=1}^{N}p_iw_i^T=0.
  \label{eq3}
\end{equation}
Then, these constraints can be rewritten in the form of matrix calculation and the parameters can be solved as follows:
\begin{equation}
    \begin{bmatrix} C\\ M \\W \end{bmatrix}=
    \begin{bmatrix} \mathbbm{1} &P& K\\ 0 &0 &\mathbbm{1}^T \\0 &0& P^T \end{bmatrix}^{-1}\begin{bmatrix} P'\\ 0 \\0 \end{bmatrix},
    \label{eq4}
 \end{equation}
where $\mathbbm{1}$ is a $N\times 1$ all-one matrix. Each element $k_{ij}$ in $K\in \mathbb{R}^{N\times N}$ is determined by $O(\parallel p_i-p_j\parallel_2 )$, and $W=[w_i,...,w_N]^T$.

Similar to the 4-pt parameterization of the homography, TPS transformation can also be parameterized as the motions of control points. In this work, we define $(U+1)\times (V+1)$ control points being evenly distributed on the target image, and then predict the motions of each control point. To bridge the global homography warp with the local TPS warp, we regress the homography transformation first to provide initial motions of control points. Then we can predict the residual motions for further flexible deformation.

\vspace{-0.3cm}
\subsubsection{Pipeline of Warp}
\label{sec:312}
\vspace{-0.1cm}

As shown in Fig.\ref{fig:network}, given $I_r$, $I_t$, we adopt ResNet50 \cite{he2016deep} with pretrained parameters as our backbone to extract semantic features first. It maps a 3-channel image to the high-dimensional semantic features with a resolution scaled to 1/16 of the original. Then the correlation between these feature maps ($F_r^{1/16}$ and $F_t^{1/16}$) can be aggregated into 2-channel feature flows using the contextual correlation layer \cite{9605632}. Subsequently, a regression network is used to estimate the 4-pt parameterization of the homography warp. This global warp also generates the initial motions of control points.

Next, we warp the feature maps with higher resolution ($F_t^{1/8}$) to embed the homographic prior into the following workflow. After another contextual correlation layer and regression network, the residual motions of control points are predicted, contributing to a robust flexible TPS warp.

\vspace{-0.3cm}
\subsubsection{Optimization of Warp}
\label{sec:313}
\vspace{-0.1cm}

To achieve content alignment and shape preservation simultaneously, we design our objective function $\mathcal{L}^{w}$ concerning two aspects: alignment and distortion.
\begin{equation}
    \mathcal{L}^{w} = \mathcal{L}_{alignment}^{w} + \omega \mathcal{L}_{distortion}^{w}.
    \label{eq5}
 \end{equation}

For the alignment, we encourage the overlapping regions to keep consistent at the pixel level.
Denoting $\varphi(\cdot,\cdot)$ is the warping operation and $\mathds{1}$ an all-one matrix with the same resolution as $I_r$, the alignment loss can be defined as follows:
\begin{equation}
\begin{aligned}
    \mathcal{L}_{alignment}^{w} = & \lambda \Vert I_{r}\cdot\varphi(\mathds{1}, \mathcal{H})- \varphi(I_{t}, \mathcal{H})\Vert_1 + \\
     &  \lambda \Vert I_{t}\cdot\varphi(\mathds{1}, \mathcal{H}^{-1})- \varphi(I_{r}, \mathcal{H}^{-1})\Vert_1 +\\
     & \Vert I_{r}\cdot\varphi(\mathds{1}, \mathcal{TPS})- \varphi(I_{t}, \mathcal{TPS})\Vert_1,
    \end{aligned}
 \end{equation}
where $\mathcal{H}$ and $\mathcal{TPS}$ are warp parameters, and $\lambda$ is a hyperparameter to balance the impacts of different transformations.

For the distortion, we link adjacent control points in the warped target image to form a mesh and introduce an inter-grid constraint $\ell_{inter}$ and an intra-grid constraint $\ell_{intra}$. The former preserves geometric structures for non-overlapping regions, while the latter reduces projective distortions.
In the beginning, we approximate a similar transformation by DLT for every grid in non-overlapping regions and take the 4-pt projective error as the loss. But this constraint that is commonly used in traditional methods \cite{he2013rectangling,liu2009content} does not work in deep learning schemes.
Instead, we re-explore the constraints from a more intuitive perspective --- the grid edge.

Similar to \cite{nie2022deep}, we penalize the grid edge $\vec{e}$ with the magnitude exceeding a threshold. Denoting $\{\vec{e}_{ hor}\}$ and $\{\vec{e}_{ver}\}$ are the collections of horizontal and vertical edges, we describe the intra-grid constraint as follows:
\begin{equation}
\begin{aligned}
  \ell_{intra}= &\frac{_1}{_{(U+1)\times V}}\sum_{\{\vec{e}_{ hor}\}}\sigma(\langle\vec{e},\vec{i}\rangle-\frac{_{2W}}{^V})
  + \\ & \frac{_1}{_{U\times (V+1)}}\sum_{\{\vec{e}_{ver}\}}\sigma(\langle\vec{e},\vec{j}\rangle-\frac{_{2H}}{^U}),
\end{aligned}
  \label{eq7}
\end{equation}
where $\vec{i}$ / $\vec{j}$ is the horizontal/vertical unit vector, and $\sigma(\cdot)$ is the $RELU$ function. The projective distortions are reduced by preventing the grid shape from dramatic scaling.

By encouraging the edge pairs (successive edges in horizontal or vertical directions, denoted as $\vec{e}_{s1},\vec{e}_{s2}$) to be co-linear, we formulate the inter-grid constraint as:
\begin{equation}
  \ell_{inter}= \frac{1}{Q}\sum_{\{\vec{e}_{s1}, \vec{e}_{s2}\}}S_{s1,s2}\cdot(1-\frac{\langle \vec{e}_{s1},\vec{e}_{s2}\rangle}{\parallel \vec{e}_{s1}\parallel \cdot \parallel \vec{e}_{s2}\parallel }),
\end{equation}
where $Q$ is the number of edge pairs and $S_{s1,s2}$ is a 0-1 label that is set to 1 if this edge pair locates on non-overlapping regions. We only preserve the structures in non-overlapping regions, preventing adverse effects on the alignment.

\subsection{Unsupervised Seamless Composition}
\label{sec:32}
\subsubsection{Motivation}
\label{sec:321}
\vspace{-0.1cm}
UDIS \cite{nie2021unsupervised} composites a stitched image via unsupervised reconstruction from feature to pixel, but it cannot deal with large parallax.
Traditional seam cutting eliminates artifacts by finding a seamless cutting path using dynamic programming \cite{avidan2007seam} or graph-cut optimization \cite{kwatra2003graphcut}, but it shows over-reliance on photometric differences.


An intuitive idea is to cooperate the motivation of seam cutting into a learning framework.
Nevertheless, how to make our unsupervised deep stitching approach work with seam cutting and be effective is a major difficulty.
For example, dynamic programming is not differential; graph-cut optimization assigns absolute integers to the labels, which truncates gradients in the backpropagation.
In this stage, we propose to relax the hard label to a \textit{soft mask} with float numbers, innovatively supervising the generation of seam-inspired masks via the balancing effect of two constraints with special designs.

\vspace{-0.3cm}
\subsubsection{Pipeline of Composition}
\label{sec:322}
\vspace{-0.1cm}
At first, we concatenate warped images as input and exploit the UNet-like network \cite{ronneberger2015u} as our composition network. But this pattern coarsely mixes the features from different images. It is challenging for such a network to perceive the semantic difference between warped images.

To overcome it, we use the encoder of the network to extract semantic features from $I_{wr}$ and $I_{wt}$ separately with shared weights. For skip connections, we replace them by subtracting the features of $I_{wt}$ from that of $I_{wr}$ and delivering the residuals at each resolution to the decoder. We set the filter number and activation function of the last layer to 1 and $sigmoid$ to predict $M_{cr}$ for the warped reference image. The other mask $M_{ct}$ for the warped target image can be easily obtained by simple post-processing.

\vspace{-0.3cm}
\subsubsection{Optimization of Composition}
\label{sec:322}
The optimization goal of our unsupervised composition includes a boundary term and a smoothness term as follows:
\begin{equation}
    \mathcal{L}^{c} = \alpha\mathcal{L}_{boundary}^{c} + \beta\mathcal{L}_{smoothness}^{c}.
 \end{equation}
The former indicates the start point and end point of the ``seam" while the latter constrains the route.

We expect the endpoints to be the intersections of the boundaries of warped images.
To achieve it, we leverage 0-1 boundary masks $M_{br}$, $M_{bt}$ to indicate the boundary positions of overlapping regions on both sides of the ``seam". \textit{More details are available in Section 3.1 of the supplementary material.} Then, we formulate the boundary loss as follows:
\begin{equation}
    \mathcal{L}_{boundary}^{c} = \parallel (S-I_{wr})\cdot M_{br}\parallel_1 + \parallel (S-I_{wt})\cdot M_{bt}\parallel_1.
 \end{equation}
This loss constrains boundary pixels of overlapping regions in $S$ from either $I_{wr}$ or $I_{wt}$. However, $M_{br}$ and $M_{bt}$ share common intersections, which produces ambiguity for the belongs of intersections. But it is the ambiguity that fixes the endpoints of a ``seam" to the intersections.

To measure the smoothness of a seam, traditional seam-cutting approaches define various energy functions with different photometric differences. In this work, we adopt the simplest photometric difference as $D=(I_{wr}-I_{wt})^2$ to demonstrate our effectiveness.
Then we define the smoothness on the difference map as follows:
\begin{equation}
\begin{aligned}
    \ell_{D} = &\sum_{i,j}|M_{cr}^{i,j}-M_{cr}^{i+1,j}|(D^{i,j}+D^{i+1,j}) +\\
    &\sum_{i,j}|M_{cr}^{i,j}-M_{cr}^{i,j+1}|(D^{i,j}+D^{i,j+1}),
     \vspace{-0.1cm}
\end{aligned}
\label{eq11}
 \end{equation}

where $i,j$ are the Cartesian coordinates. To produce a smooth transition between both sides of the ``seam", we also define the smoothness of the stitched image as follows:
\begin{equation}
\begin{aligned}
    \ell_{S} = &\sum_{i,j}|M_{cr}^{i,j}-M_{cr}^{i+1,j}|\cdot|S^{i,j}-S^{i+1,j}| +\\
    &\sum_{i,j}|M_{cr}^{i,j}-M_{cr}^{i,j+1}|\cdot|S^{i,j}-S^{i,j+1}|.
      \vspace{-0.1cm}
\end{aligned}
\label{eq12}
 \end{equation}

By adding $\ell_{D}$ and $\ell_{S}$, we formulate the complete smoothness term $\mathcal{L}_{smoothness}^{c}$.
Note that, our network is trained to facilitate the capability to extract semantic differences. In the inference process, the proposed method no longer relies on photometric differences.

\begin{figure}[!t]
  \centering
  \includegraphics[width=0.44\textwidth]{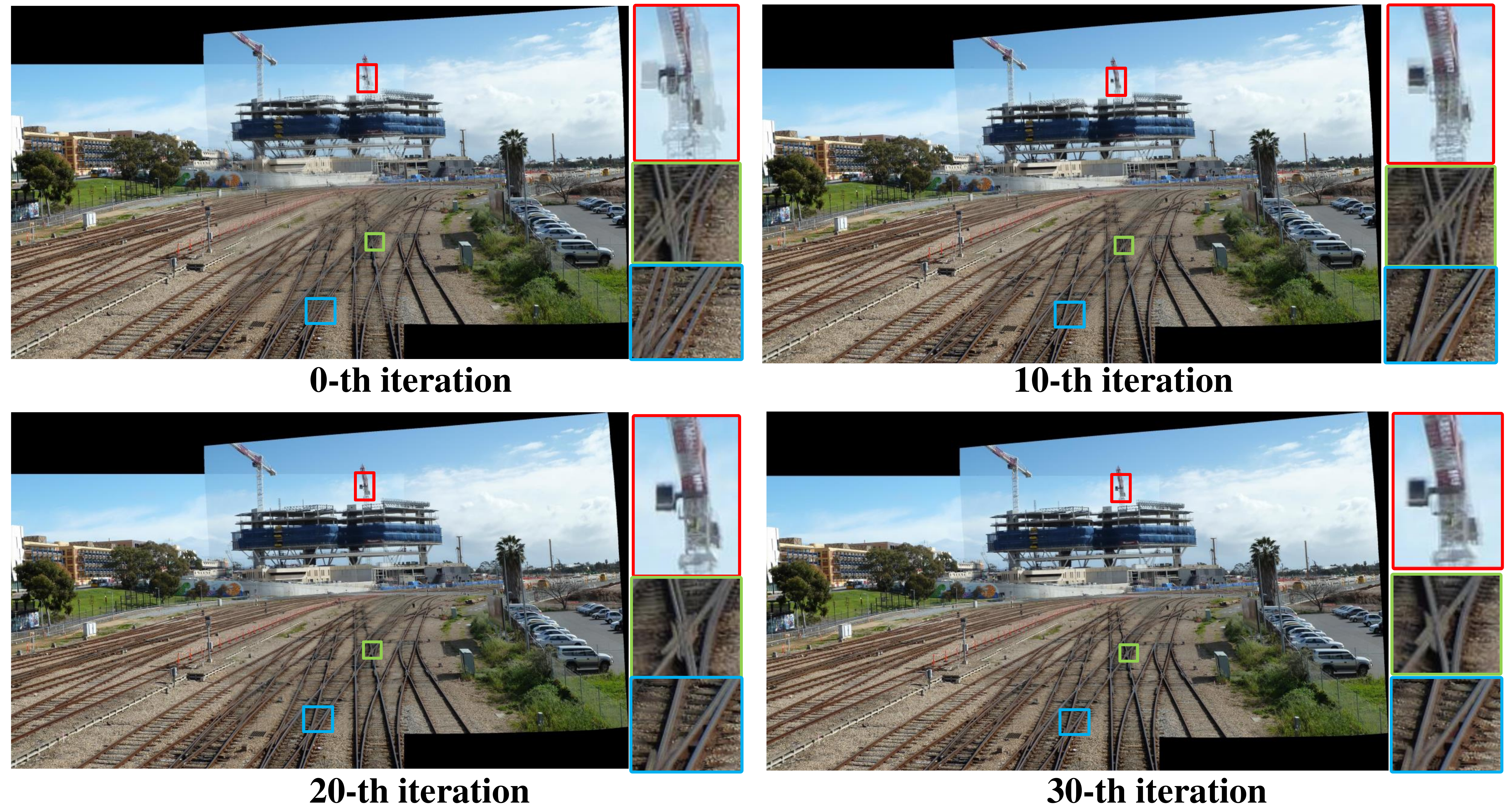}
  \vspace{-4pt}
  \caption{We demonstrate the process of iterative warp adaption on ``railtrack" dataset \cite{zaragoza2013projective} (cross-dataset and cross-resolution).}
  \label{fig:iter}
  \vspace{-0.3cm}
\end{figure}

\subsection{Iterative Warp Adaption}
\label{sec:33}

To transfer a pretrained model to other datasets (cross-scene and cross-resolution), the most common way is to fine-tune on the new dataset.
However, it usually requires labels to assist the adaption process. In this work, we address this limitation by setting an unsupervised optimization goal as follows:
\begin{equation}
    \mathcal{L}_{adaption}^{w} = \Vert I_{r}\cdot\varphi(\mathds{1}, \mathcal{TPS})- \varphi(I_{t}, \mathcal{TPS})\Vert_1.
    \label{eq13}
\end{equation}
Compared with Eq. \ref{eq5}, we remove the homography alignment loss and distortion loss. Because these constraints have been well learned by the pretrained model, what we do is adjust the local alignment on different data.

Furthermore, we consider a special case that the new dataset only contains one sample. Experiments exhibit that our model can also be optimized stably for adapting to only sample in an iterative fashion. In particular, we set a threshold $\tau$ and a maximum iteration number $T$. The adaption process stops when the iteration number reaches $T$ or consecutive optimization errors (Eq. \ref{eq13}) are lower than $\tau$.



We show an iterative adaption example in Fig. \ref{fig:iter}, where the artifacts are significantly reduced with the increase of iteration number. It takes about 0.1s to finish an iteration.
\vspace{-0.1cm}
\section{Experiments}
\label{sec:experiment}
\vspace{-0.1cm}
\subsection{Dataset and Implement Details}
\vspace{-0.1cm}
\label{sec:41}
\textit{Dataset:} To make an intuitive and fair comparison with deep stitching methods, we also train our model on UDIS-D \cite{nie2021unsupervised} dataset. The evaluation is conducted on UDIS-D dataset and other traditional datasets \cite{zaragoza2013projective, gao2011constructing, lin2015adaptive, li2017parallax, lin2016seagull}.

\vspace{0.cm}
\textit{Details:} We train our warp and composition networks for 100 and 50 epochs using Adam \cite{kingma2014adam} with an exponentially decaying learning rate with an initial value of $10^{-4}$. For the warp stage, $\omega$ and $\lambda$ are set to 10 and 3, and we adopt $(12+1)\times (12+1)$ control points to provide the flexible TPS transformation. For the second stage, we set $\alpha$ and $\beta$ to 10,000 and 1,000. As for the warp adaption, $\tau$ and $T$ are assigned as $10^{-4}$ and 50. All implementations are based on PyTorch using a single GPU with NVIDIA RTX 3090 Ti.
\vspace{-0.3cm}

\begin{table*}[!t]
  \centering
  \caption{Quantitative comparison of warp on UDIS-D dataset \cite{nie2021unsupervised}. The best is marked in red and the second best is in blue.}
  \vspace{-0.1cm}
  \scalebox{0.92}{
  \begin{tabular}{lllll|llll}
   \toprule
    &\multicolumn{4}{c}{PSNR $\uparrow$}&\multicolumn{4}{c}{SSIM $\uparrow$}\\
   \cline{2-9}
   & \makecell[c]{Easy} & \makecell[c]{Moderate}& \makecell[c]{Hard}& \makecell[c]{Average}  & \makecell[c]{Easy} & \makecell[c]{Moderate}& \makecell[c]{Hard}& \makecell[c]{Average}\\
   \midrule
  \makecell[c]{$I_{3\times 3}$} & \makecell[c]{15.87} & \makecell[c]{12.76}& \makecell[c]{10.68}& \makecell[c]{12.86}  & \makecell[c]{0.530} & \makecell[c]{0.286}& \makecell[c]{0.146}& \makecell[c]{0.303}\\
  \makecell[c]{SIFT\cite{lowe2004distinctive}+RANSAC\cite{fischler1981random}} & \makecell[c]{28.75} & \makecell[c]{24.08}& \makecell[c]{18.55}& \makecell[c]{23.27}  & \makecell[c]{0.916} & \makecell[c]{0.833}& \makecell[c]{0.636}& \makecell[c]{0.779}\\
  \makecell[c]{APAP\cite{zaragoza2013projective}} & \makecell[c]{27.96} & \makecell[c]{24.39}& \makecell[c]{$\mathbf{\textcolor{blue}{20.21}}$}& \makecell[c]{23.79}  & \makecell[c]{0.901} & \makecell[c]{0.837}& \makecell[c]{0.682}& \makecell[c]{0.794}\\
  \makecell[c]{ELA\cite{li2017parallax}} & \makecell[c]{$\mathbf{\textcolor{blue}{29.36}}$} & \makecell[c]{$\mathbf{\textcolor{blue}{25.10}}$}& \makecell[c]{19.19}& \makecell[c]{$\mathbf{\textcolor{blue}{24.01}}$}  & \makecell[c]{$\mathbf{\textcolor{blue}{0.917}}$} & \makecell[c]{$\mathbf{\textcolor{blue}{0.855}}$}& \makecell[c]{$\mathbf{\textcolor{blue}{0.691}}$}& \makecell[c]{$\mathbf{\textcolor{blue}{0.808}}$}\\
  \makecell[c]{SPW\cite{liao2019single}} & \makecell[c]{26.98} & \makecell[c]{22.67}& \makecell[c]{16.77}& \makecell[c]{21.60}  & \makecell[c]{0.880} & \makecell[c]{0.758}& \makecell[c]{0.490}& \makecell[c]{0.687}\\
  \makecell[c]{LPC\cite{jia2021leveraging}} & \makecell[c]{26.94} & \makecell[c]{22.63}& \makecell[c]{19.31}& \makecell[c]{22.59}  & \makecell[c]{0.878} & \makecell[c]{0.764}& \makecell[c]{0.610}& \makecell[c]{0.736}\\
  \makecell[c]{UDIS's warp\cite{nie2021unsupervised}} & \makecell[c]{25.16} & \makecell[c]{20.96}& \makecell[c]{18.36}& \makecell[c]{21.17}  & \makecell[c]{0.834} & \makecell[c]{0.669}& \makecell[c]{0.495}& \makecell[c]{0.648}\\
  \makecell[c]{Our warp} & \makecell[c]{$\mathbf{\textcolor{red}{30.19}}$} & \makecell[c]{$\mathbf{\textcolor{red}{25.84}}$}& \makecell[c]{$\mathbf{\textcolor{red}{21.57}}$}& \makecell[c]{$\mathbf{\textcolor{red}{25.43}}$}  & \makecell[c]{$\mathbf{\textcolor{red}{0.933}}$} & \makecell[c]{$\mathbf{\textcolor{red}{0.875}}$}& \makecell[c]{$\mathbf{\textcolor{red}{0.739}}$}& \makecell[c]{$\mathbf{\textcolor{red}{0.838}}$}\\

      \bottomrule
   \end{tabular}
  }
   \vspace{-0.1cm}
   \label{tab1}
   \end{table*}
\vspace{-0.cm}

\begin{figure*}[!t]
  \centering
  \includegraphics[width=0.92\textwidth]{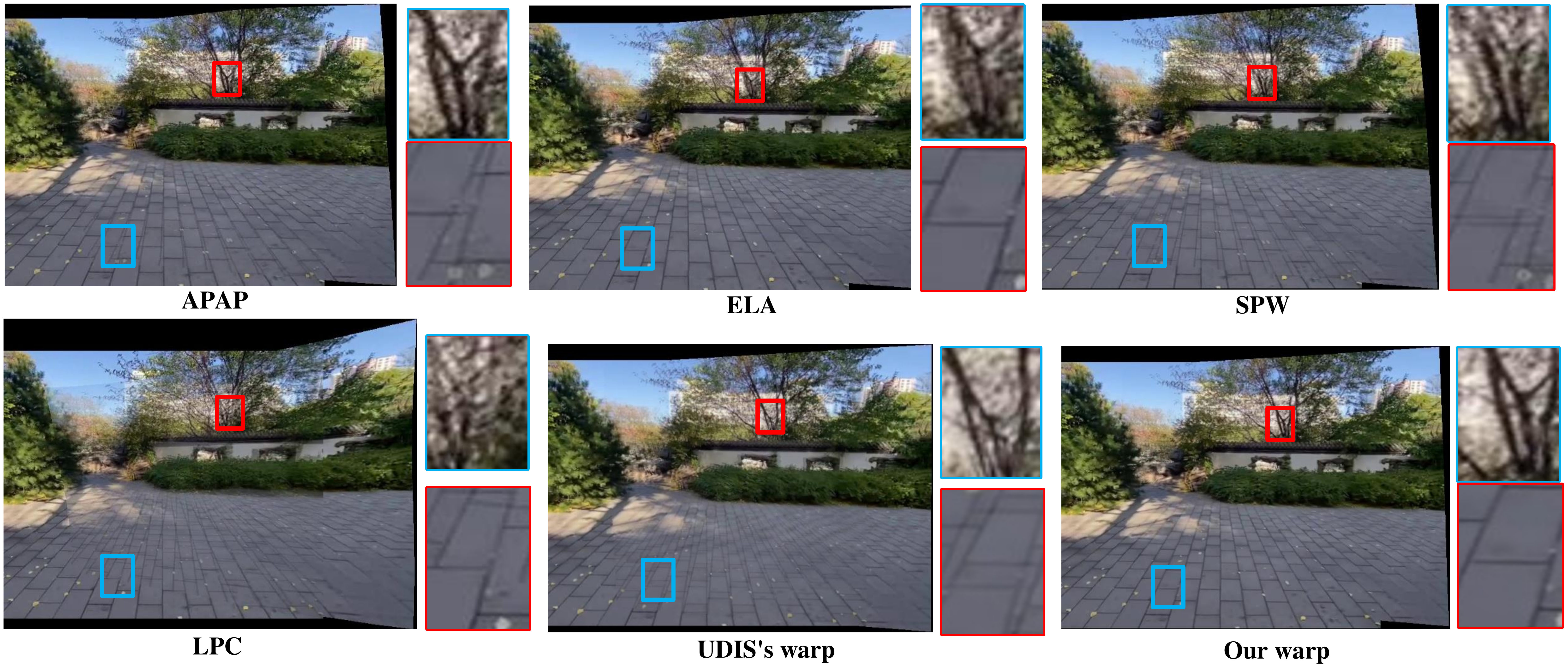}
  \vspace{-4pt}
  \caption{Qualitative comparison of warp on UDIS-D dataset \cite{nie2021unsupervised}. We zoom in on a near region and a far region to show the alignment performance. \textit{For clarity, we show the inputs and more comparative results in the supplementary material.}}
  \label{fig:warp1}
\end{figure*}

\begin{figure*}[!t]
  \centering
  \includegraphics[width=0.93\textwidth, height=0.35\textheight]{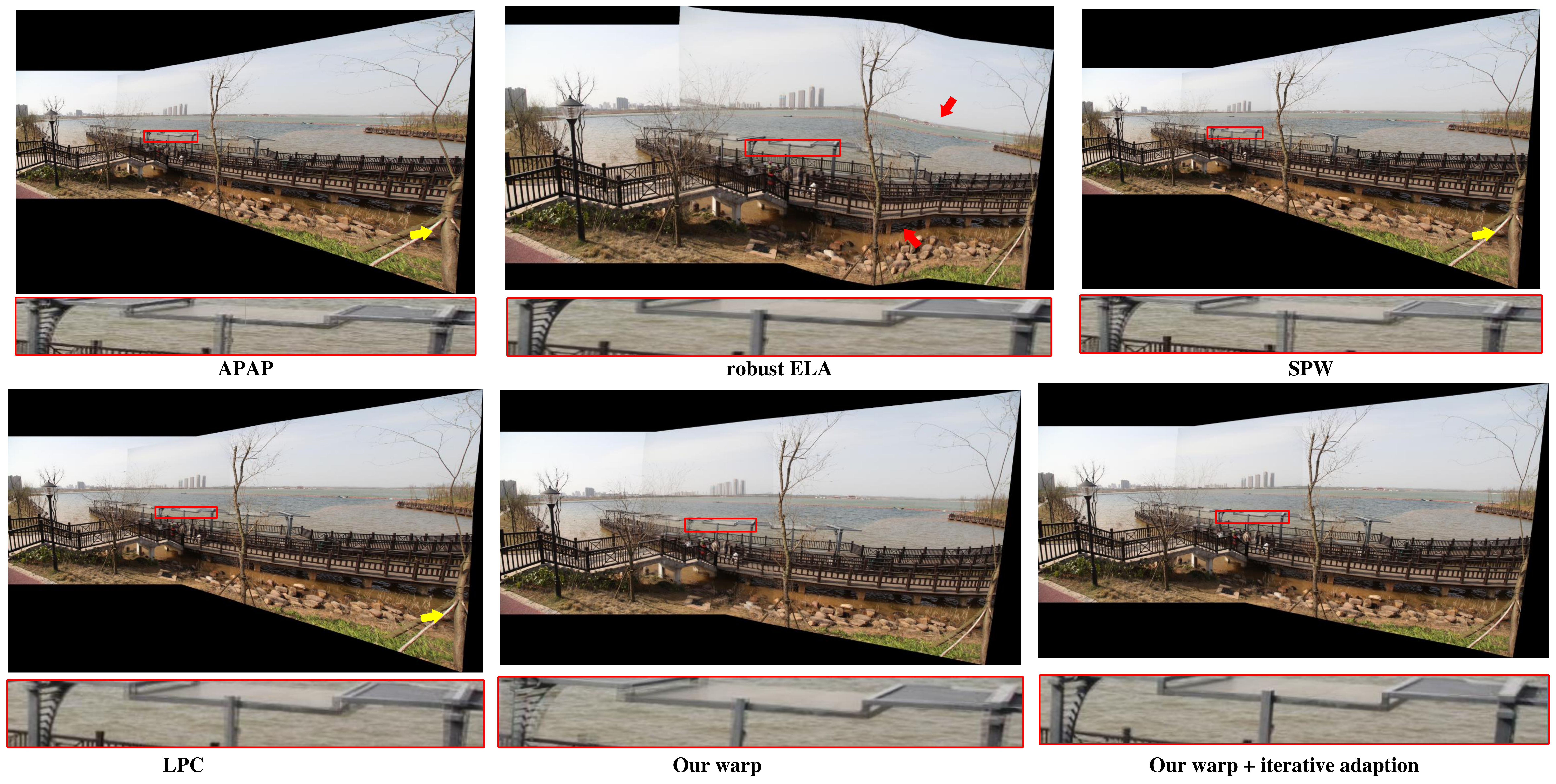}
  \vspace{-4pt}
  \caption{Qualitative comparison of warp on ``boardingBridge" dataset \cite{li2017parallax} with a resolution of $1440\times 2160$ for inputs. The yellow and red arrows highlight projective and structural distortions. }
  \label{fig:warp2}
  \vspace{-0.3cm}
\end{figure*}

\begin{table}[!t]
  \centering
  \caption{Comparison of warp on elapsed time (s). 1: tested with Intel i7-9750H 2.60GHz CPU; 2: tested with NVIDIA RTX 3090Ti GPU.}
  \vspace{-0.2cm}
  \scalebox{0.82}{
  \begin{tabular}{llll}
   \toprule
    \makecell[c]{Dataset} & \makecell[c]{Railtrack \cite{zaragoza2013projective}}&\makecell[c]{Fence \cite{lin2015adaptive}}&\makecell[c]{Carpark \cite{gao2011constructing}}\\
   \cline{1-4}
   \makecell[c]{Resolution} & \makecell[c]{$1500\times 2000$} & \makecell[c]{$1088\times 816$} & \makecell[c]{$490\times 653$}\\
   \midrule
   \makecell[c]{APAP \cite{zaragoza2013projective}\textsuperscript{1}} & \makecell[c]{20.921} & \makecell[c]{4.427} & \makecell[c]{2.005}\\
   \makecell[c]{ELA \cite{li2017parallax}\textsuperscript{1}} & \makecell[c]{18.982} & \makecell[c]{4.739} & \makecell[c]{2.179}\\
   \makecell[c]{SPW \cite{liao2019single}\textsuperscript{1}} & \makecell[c]{227.762} & \makecell[c]{4.787} & \makecell[c]{6.583}\\
   \makecell[c]{LPC \cite{jia2021leveraging}\textsuperscript{1}} & \makecell[c]{2805.3} & \makecell[c]{9.115} & \makecell[c]{40.443}\\
   \makecell[c]{Our warp\textsuperscript{1}} & \makecell[c]{12.073} & \makecell[c]{5.025} & \makecell[c]{3.486}\\
   \makecell[c]{Our warp\textsuperscript{2}} & \makecell[c]{\textbf{0.731}} & \makecell[c]{\textbf{0.210}} & \makecell[c]{\textbf{0.117}}\\

      \bottomrule
   \end{tabular}
  }
  \vspace{-0.1cm}

   \label{tab2}
   \end{table}

\begin{table}[!t]
  \centering
  \caption{Comparison of composition on elapsed time (s). 1: tested with Intel i7-9750H 2.60GHz CPU; 2: tested with NVIDIA RTX 3090Ti GPU.}
  \vspace{-0.2cm}
  \scalebox{0.82}{
  \begin{tabular}{llll}
   \toprule
    \makecell[c]{Dataset} & \makecell[c]{Railtrack \cite{zaragoza2013projective}}&\makecell[c]{Fence \cite{lin2015adaptive}}&\makecell[c]{Carpark \cite{gao2011constructing}}\\
   \cline{1-4}
   \makecell[c]{Resolution\\(after warping)} & \makecell[c]{$1831\times 3193$} & \makecell[c]{$1298\times 1320$} & \makecell[c]{$718\times 1186$}\\
   \midrule
   \makecell[c]{Seam cutting \cite{li2018perception}\textsuperscript{1}} & \makecell[c]{46.657} & \makecell[c]{4.058} & \makecell[c]{0.873}\\
   \makecell[c]{Reconstruction \cite{nie2021unsupervised}\textsuperscript{1}} & \makecell[c]{304.963} & \makecell[c]{80.837} & \makecell[c]{10.734}\\
   \makecell[c]{Our composition\textsuperscript{1}} & \makecell[c]{22.778} & \makecell[c]{6.666} & \makecell[c]{3.286}\\
   \makecell[c]{Our composition\textsuperscript{2}} & \makecell[c]{\textbf{0.532}} & \makecell[c]{\textbf{0.143}} & \makecell[c]{\textbf{0.071}}\\

      \bottomrule
   \end{tabular}
  }
  \vspace{-0.4cm}
   \label{tab3}
\end{table}



\begin{figure*}[!t]
  \centering
  \includegraphics[width=0.92\textwidth]{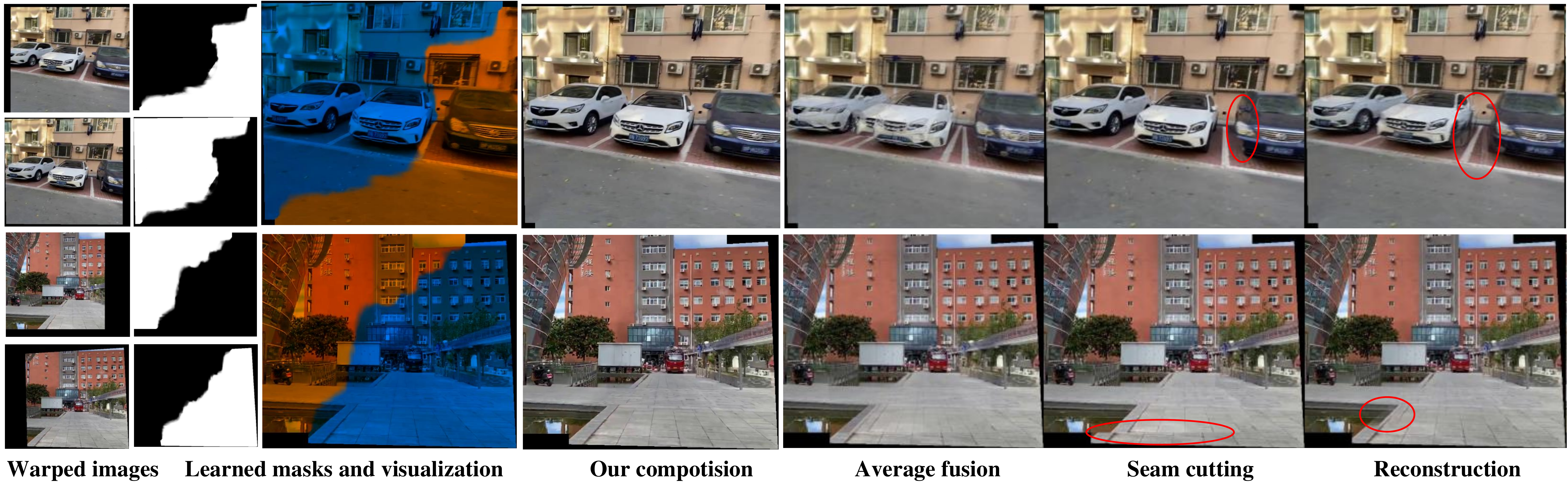}
  \vspace{-4pt}
  \caption{The comparison of composition. \textit{For clarity, more results are reported in the supplementary material.}}
  \label{fig:composition}
  \vspace{-0.2cm}
\end{figure*}
\vspace{5pt}

\begin{figure*}[!t]
  \centering
  \includegraphics[width=0.92\textwidth, height=0.26\textheight]{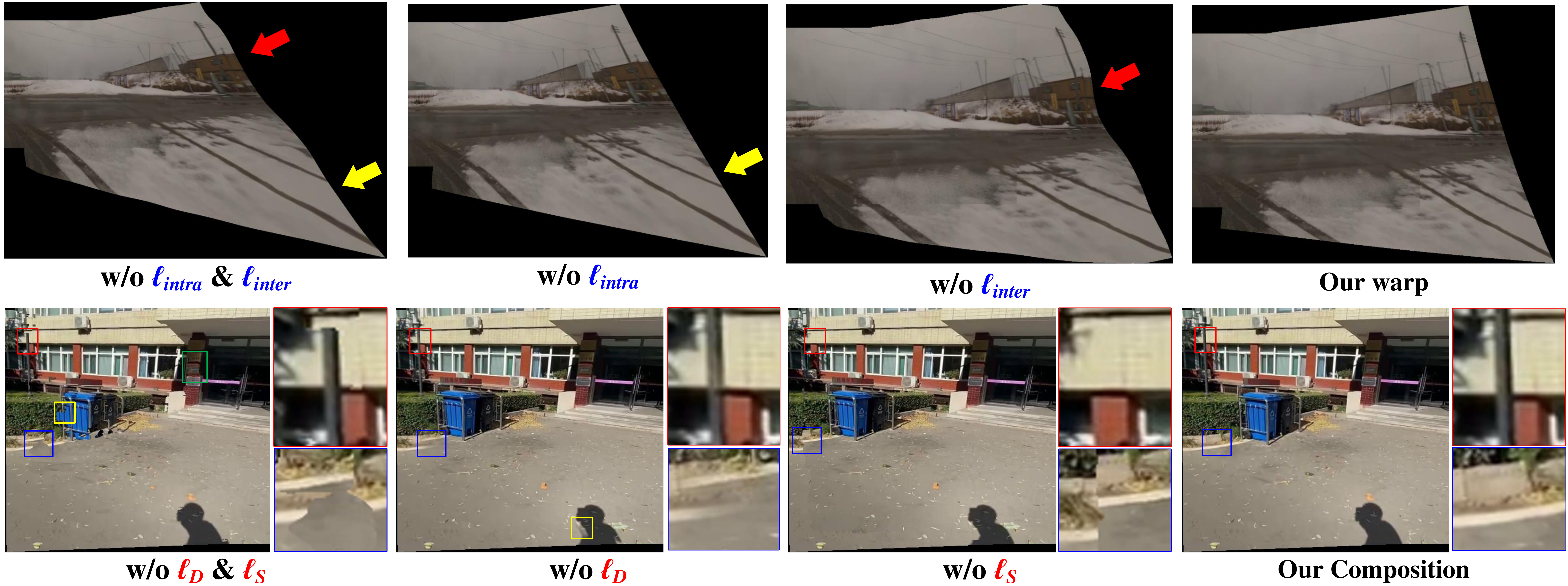}
  \vspace{-4pt}
  \caption{Ablation studies on our warp and composition. Top: the red and yellow arrows highlight the structural and projective distortions, respectively. Bottom: the rectangles indicate the discontinuous regions.}
  \label{fig:ablation}
  \vspace{-0.3cm}
\end{figure*}

\subsection{Comparative Experiments}
\vspace{-0.1cm}
\label{sec:42}
To demonstrate our effectiveness comprehensively, we conduct extensive experiments on warp, composition, and the complete stitching framework, respectively.

\vspace{-0.4cm}
\subsubsection{Comparisons of Warp}
\label{sec:421}
\vspace{-0.1cm}

We compare our warp with SIFT \cite{lowe2004distinctive}+RANSAC \cite{fischler1981random} (the pipeline of AutoStitch \cite{brown2007automatic}), APAP \cite{zaragoza2013projective}, ELA \cite{li2017parallax}, SPW \cite{liao2019single}, LPC \cite{jia2021leveraging}, and UDIS's warp \cite{nie2021unsupervised}. We implement SIFT+RANSAC by ourselves and adopt the official codes for other methods with default parameters such as mesh resolutions. All the methods, including ours, use the average fusion as the post-processing operation. Because this simple fusion is fast and can better highlight the misalignments.

\vspace{0.0cm}
\textit{Quantitative comparison:} We first carry on quantitative comparisons with the same metrics as UDIS \cite{nie2021unsupervised} on UDIS-D dataset \cite{nie2021unsupervised} that has 1,106 samples for the evaluation. The results are shown in Table \ref{tab1}, where $I_{3\times 3}$ takes the identity matrix as a ``no-warping" transformation for reference. The results are divided into three parts according to the performance as \cite{nie2021unsupervised, 9605632}. The programs of traditional methods might crash in some challenging samples due to the lack of geometric features. When that happens, we use $I_{3\times 3}$ as an alternative transformation for the evaluation.

\vspace{0.0cm}
\textit{Qualitative comparison:} Qualitative results are shown in Fig. \ref{fig:warp1}, where we zoom in on two regions at different depth surfaces to highlight parallax artifacts. From this figure, our warp outperforms the other solutions by a large margin on UDIS-D dataset \cite{nie2021unsupervised}.

\vspace{0.0cm}
\textit{Cross-dataset comparison:} We use the pretrained model to evaluate our performance on other datasets, as illustrated in Fig. \ref{fig:warp2}. The iterative adaption strategy is used to further improve the alignment performance. 

\vspace{0.0cm}
\textit{Speed comparison:} To evaluate the speed objectively, we test it on three traditional public datasets \cite{zaragoza2013projective, lin2015adaptive, gao2011constructing} with three different resolutions. As reported in Table \ref{tab2}, our warp has a speed far exceeding the others with GPU acceleration, while traditional warps cannot be accelerated by GPU. For traditional mesh-based warps, the runtime does not vary linearly with the resolution, and in scenes with rich geometric features (e.g., ``railTrack"), the speed becomes a disaster.

\vspace{-0.3cm}
\subsubsection{Comparisons of Composition}
\label{sec:422}
\vspace{-0.2cm}

We compare our composition with the perception-based seam-cutting approach \cite{li2018perception} and reconstruction-based method \cite{nie2021unsupervised}. To show the parallax artifacts more intuitively, we warp the images by SIFT+RANSAC and give the results of average fusion for reference.

\vspace{0.0cm}
\textit{Qualitative comparison:} Traditional seam-cutting methods find the seam by dynamic programming \cite{avidan2007seam} or graph-cut optimization \cite{kwatra2003graphcut}. The values in traditional masks are integers while that in ours are float. Therefore, we cannot evaluate our composition quantitatively with traditional indicators.
Instead, we show qualitative results in Fig. \ref{fig:composition}. Besides, we promise to release all subjective results, including 1,106 images in UDIS-D and others in traditional datasets. 

\vspace{0.0cm}
\textit{Speed comparison:} Here, we warp the inputs with the proposed warp first. Then these warped images are used for speed evaluation on different composition methods. As illustrated in Table. \ref{tab3}, our composition shows significant speed superiority over the others with GPU acceleration.

\vspace{-0.3cm}
\subsubsection{More Comparisons}
\label{sec:423}
\vspace{-0.2cm}
Here, we evaluate the performance of our complete stitching framework with other SoTA methods. The results are illustrated in Fig. \ref{fig1}, where LPC \cite{jia2021leveraging} and UDIS \cite{nie2021unsupervised} adopt the perception-based seam cutting \cite{li2018perception} and reconstruction \cite{nie2021unsupervised} for the post-processing operations.
\textit{For clarity, more experimental results including qualitative comparisons, user studies, challenging cases, and cross-dataset evaluations are depicted in the supplementary material.}

\subsection{Ablation studies}
\label{sec:43}
\vspace{-0.1cm}
We first conduct ablation studies on different warp constraints. As shown in Fig. \ref{fig:ablation}(top), the inter-gird constraint preserves the structures whiles the intra-grid one reduces projective distortions. Moreover, these constraints bring little adverse impact on alignment. \textit{Quantitative results are reported in the supplementary material.}

Then we study the impacts of smoothness term in our composition. The results are shown in Fig. \ref{fig:ablation}(bottom), where we highlight the discontinuous regions by rectangles. With the smoothness constraints on the difference map and stitched image, the discontinuity is significantly improved.



\vspace{-0.4cm}
\section{Conclusion}
\label{sec:conclusion}
\vspace{-0.1cm}
In this paper, we propose a parallax-tolerant unsupervised deep stitching solution. First, a robust flexible warp is adaptively learned for both content alignment and shape preservation. We also present the seam-inspired composition to further reduce artifacts. Besides, a simple iterative warp adaption strategy is designed to effectively enhance the generalization in cross-dataset and cross-resolution cases.
Compared with existing solutions, our method can address both challenging scenes and large-parallax cases. With increasingly popular GPUs, our solution exhibits incredible efficiency.

{\small
\bibliographystyle{ieee_fullname}
\bibliography{egbib}
}

\clearpage
\appendix

\section{Supplemental Material}
\label{sec:1}
In this document, we provide the following supplementary contents:
\begin{itemize}
    \item Details of warp (Section \ref{detail_warp}).
    \item Details of composition (Section \ref{detail_composition}).
    \item Analysis on robustness and distortion (Section \ref{analysis}).
    \item More results (Section \ref{results}).
 \end{itemize}

Regarding the network architecture, we have not provided specific details such as layers, channels, etc., as we would like readers to focus more on the motivations behind our approach to solving the problem. For the details, we promise to release the code for reference.

\section{More Details of Warp}
\label{detail_warp}
\subsection{Physicality of TPS}
\label{Physicality_tps}
The thin plate spline (TPS) method can simulate arbitrary 2D deformation through the use of a deformable thin plate, which is more general than using homography. When all control points are correctly matched, we aim to use a thin plate with minimal curvatures. We then formulate an energy optimization problem that involves both alignment and distortion, as described in \cite{bookstein1989principal}:
\begin{equation}
\label{supp::eq1}
  \varepsilon	 = \varepsilon_{alignment} + \lambda \varepsilon_{distortion},
 \end{equation}
where $\lambda$ is a balancing factor to control the smoothness of the warp. The alignment energy and distortion energy are defined as follows:
\begin{equation}
\begin{aligned}
    \varepsilon_{alignment} = &\sum_{i=1}^{N}\parallel p'-\mathscr{T}(p)\parallel^2,\\
    \varepsilon_{distortion} = & \iint_{\mathbb{R}^{2}}\left(\left(\frac{\partial^{2} \mathscr{T}}{\partial x^{2}}\right)^{2}\right.+2\left(\frac{\partial^{2} \mathscr{T}}{\partial x \partial y}\right)^{2} \\
    &+\left. \left(\frac{\partial^{2} \mathscr{T}}{\partial y^{2}}\right)^{2}\right) dx dy,
    \end{aligned}
 \end{equation}
where $\mathscr{T}(\cdot)$ is the warp function.
When $\lambda>0$, the control points are allowed to be slightly misaligned in order to produce a warp with less distortion. However, in our implementation, we set $\lambda=0$ to strongly constrain the motion of the control points. This means that our network predicts the motions of the control points, and enforces the real motions ($\mathscr{T}(p)-p$) to be equal to the predicted motions. By minimizing Eq. \ref{supp::eq1}, we are able to determine the warp function, which is derived as follows (see Eq. \textcolor{red}{2} in the manuscript):
\vspace{-0.4cm}
\begin{equation}
  p'=\mathscr{T}(p)=C+Mp+\sum_{i=1}^{N}w_iO(\parallel p-p_i\parallel_2 ).
  \label{tps}
\end{equation}

\vspace{-0.5cm}
\subsection{Discussion of Alignment and Distortion}
\label{discussion}
In the previous section, we explained that aligning all control points causes distortion in the warp function. To mitigate this issue, we avoid increasing $\lambda$ in Eq. \ref{supp::eq1} and instead assume that control points are evenly distributed in the target image, and their motion is smooth. We form a mesh by connecting control points and introduce an intra-grid constraint (Eq. \textcolor{red}{7} of the manuscript) and an inter-grid constraint (Eq. \textcolor{red}{8} of the manuscript) for content preservation.

To summarize, the proposed warp yields two improvements. (1) Our network architecture with TPS benefits the alignment in overlapping regions. (2) The distortion loss (Eq. \textcolor{red}{7,8} of the manuscript) benefits the distortion elimination in non-overlapping regions.

\label{sec:3}
\begin{figure}[!t]
  \centering
  \includegraphics[width=0.48\textwidth, height=0.15\textheight]{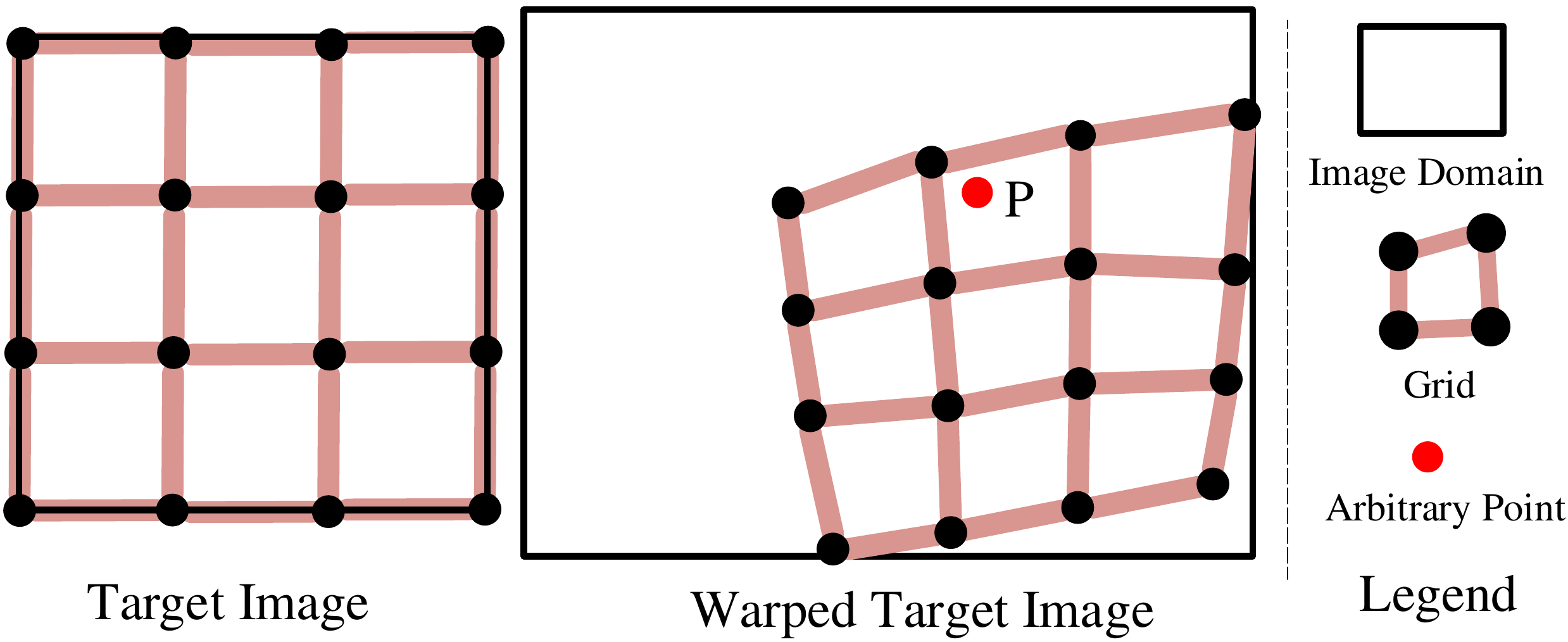}
  \vspace{-0.7cm}
  \caption{Backward interpolation.}
  \label{Interpolation}
  \vspace{-0.5cm}
\end{figure}

\subsection{Multiple Homography vs. TPS}
\label{vs}
The TPS warp is more appropriate than the traditional mesh-based multi-homography warp \cite{zaragoza2013projective} in deep stitching. Here, we discuss the reason in detail.

The multi-homography stitching methods warp the target image into a warped target image through mesh deformation as illustrated in Fig. \ref{Interpolation}. In the implementation, backward interpolation is commonly leveraged to avoid invalid pixels like holes. In backward interpolation, for an arbitrary point $P$ in a warped target image, we need to calculate the corresponding location in the target image. Then bilinear interpolation is leveraged to obtain the pixel value of $P$. Therefore, how to calculate the corresponding position is the key problem. To make it, the first thing is to determine which grid dose $P$ belong to in multi-homography warp. In the case of Fig. \ref{Interpolation}, it seems easy to find that $P$ belongs to the second grid, so we could calculate the corresponding homography through the four pairs of vertices of this grid. \textit{However, how to determine the belongings of all points in the warped target image in an efficient parallel manner makes a big difficulty.} Because the warped mesh has an irregular shape, in which even the non-convex grid might be produced. This process is hard to be parallelly accelerated, especially in GPUs, making the training time unbearable. (Empirically, the training process might take millions of iterations.)

In contrast, the TPS transformation has the advantage that all pixels share the same warp function (Eq. \ref{tps}), eliminating the need to determine the belonging of each pixel to a particular grid. In the multi-homography scheme, the warp of a pixel is determined by only four pairs of vertices, while in TPS, it is influenced by all pairs of control points ($(U+1)\times (V+1)$ in our paper). As a result, the backward interpolation of all pixels in the warped target image can be efficiently achieved in a parallel manner for TPS, making the training process faster compared to multi-homography.

\subsection{Difference to Stitching Methods using TPS}
The existing stitching methods using TPS are all traditional feature-based solutions. For example, ELA \cite{li2017parallax} calculates TPS transformations using matched keypoints such as SIFT. This transformation is then processed to reduce computational cost and distortions.

In contrast, the proposed method is the first deep learning-based stitching scheme that utilizes TPS transformations. The calculation of this warp is no longer reliant on matched keypoints. Instead, we initially define control points that are evenly distributed in the target image and then predict the motions of these points using the unsupervised network. Through the initial control points and the predicted motions, we obtain two sets of control points with one-to-one correspondence. We then formulate the warp and eliminate projective and structural distortions using intra-grid and inter-grid constraints as additional loss functions. Compared to ELA, our proposed method achieves superior alignment (Table \textcolor{red}{1} of the manuscript), fewer distortions (Fig. \textcolor{red}{5} of the manuscript), and better efficiency (Table \textcolor{red}{2} of the manuscript).

\begin{figure}[t]
   \begin{center}
      \subfloat[Left: a composition case. Right: the legend.]{\includegraphics[width=0.43\textwidth]{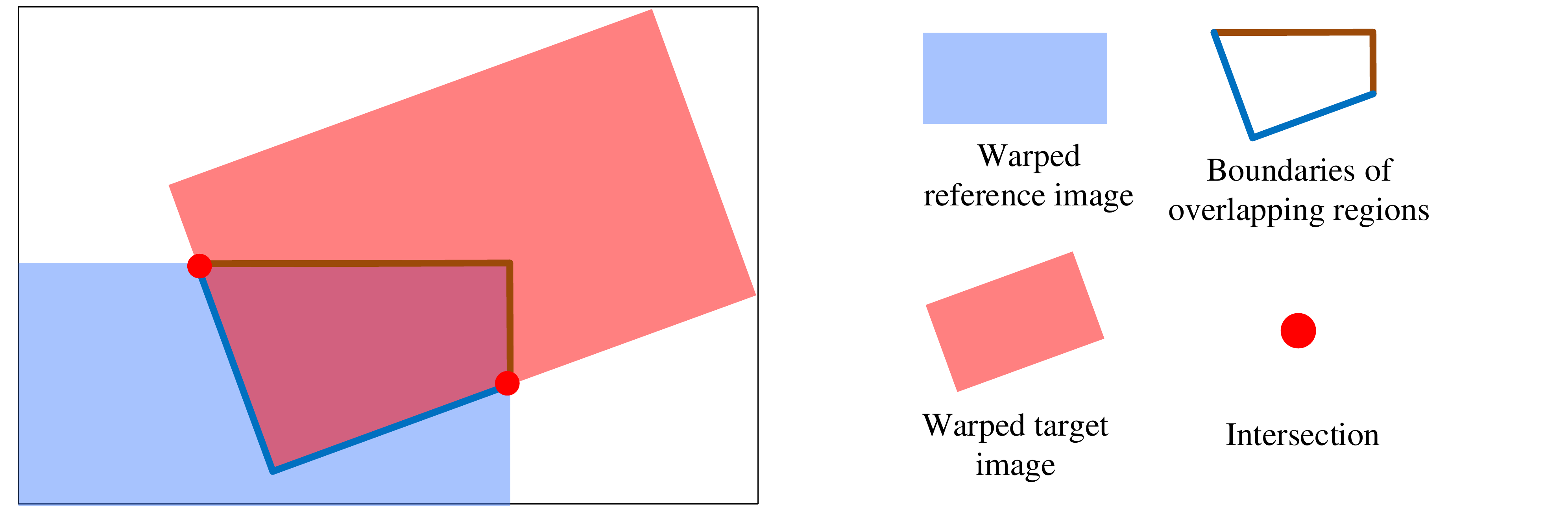}}
      \vspace{-0.3cm}
      \quad
      \subfloat[Boundary masks: $M_{br}$ (left) and $M_{bt}$ (right).]{\includegraphics[width=0.43\textwidth]{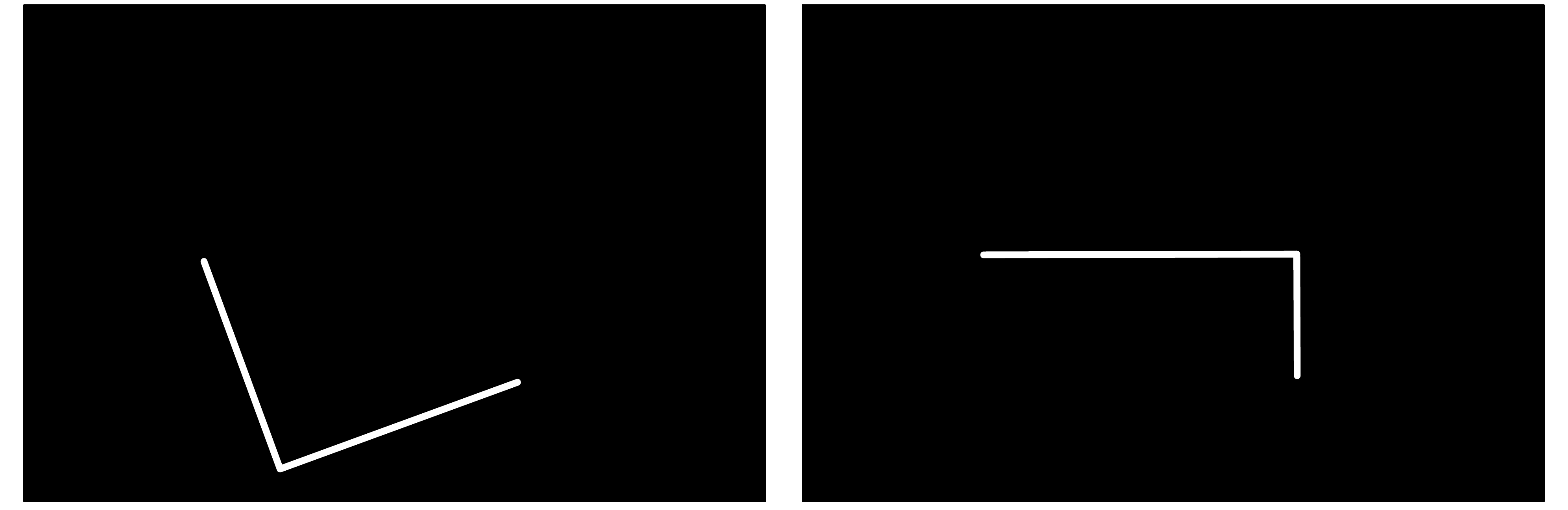}}
      \vspace{-0.3cm}
      \quad
      \subfloat[Warped masks: $M_{r}$ (left) and $M_{t}$ (right).]{\includegraphics[width=0.43\textwidth]{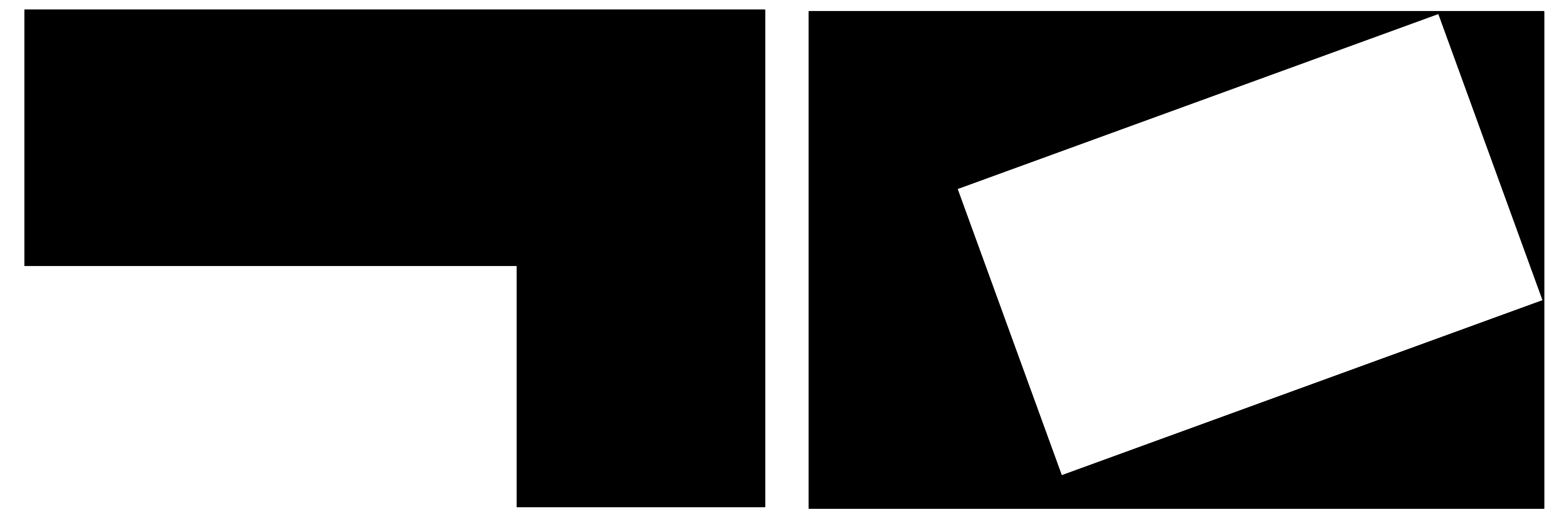}}
     \vspace{-0.5cm}
   \end{center}
   \caption{Details of the boundary term for the composition.}
   \vspace{-0.5cm}
\label{fig1}
\end{figure}


\begin{figure*}[!t]
  \centering
  \includegraphics[width=0.95\textwidth]{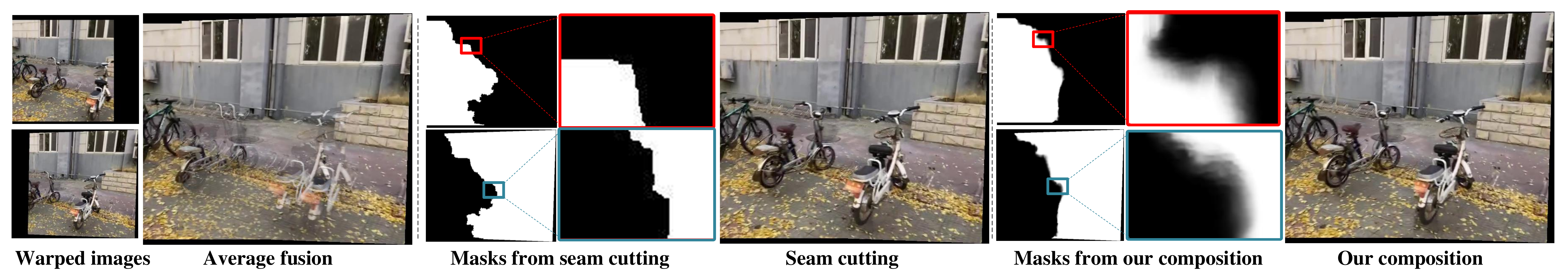}
  \vspace{-0.3cm}
  \caption{The difference between masks from seam cutting \cite{li2018perception} and our composition.}
  \label{comp_diff}
  \vspace{-0.3cm}
\end{figure*}

\section{More Details of Composition}
\label{detail_composition}
\subsection{Boundary Term}
\label{boundary_term}
Considering a composite case (Fig. \ref{fig1}\textcolor{red}{a}), we aim to fix the endpoints of a seam on the intersections. To achieve this, we define two boundary masks, as shown in Fig. \ref{fig1}\textcolor{red}{b}: $M_{br}$ and $M_{bt}$. The two boundaries are located inside the warped reference image and the warped target image, respectively. In our boundary constraint, we encourage the boundary pixels of overlapping regions in $S$ to be from either $I_{wr}$ or $I_{wt}$ using the following equation:
\begin{equation}
    \mathcal{L}_{boundary}^{c} = \parallel (S-I_{wr})\cdot M_{br}\parallel_1 + \parallel (S-I_{wt})\cdot M_{bt}\parallel_1.
 \end{equation}
By constraining the values of boundary pixels in a stitched image, we constrain that in composition masks indirectly. More importantly, $M_{br}$ and $M_{bt}$ share two common intersections as represented by the red circles in Fig.\ref{fig1}\textcolor{red}{a}. These common pixels inevitably yield ambiguity for the belongs of intersections, and the ambiguity helps to determine the seam endpoints.

Next, we describe how to get the boundary masks. Given the warped masks $M_{r}$, $M_{t}$ (as shown in Fig. \ref{fig1}\textcolor{red}{c}), we obtain boundary masks by the following formulation:
\begin{equation}
\begin{aligned}
    M_{br} = M_{r}\cdot \mathscr{E}(M_{t}),\\
    M_{bt} = M_{t}\cdot \mathscr{E}(M_{r}),
    \end{aligned}
 \end{equation}
where $\mathscr{E}(\cdot)$ denotes the edge extraction operation that can be implemented by several convolutional layers with $SOBEL$ filters.

\subsection{Difference to Seam Cutting}
Traditional seam-cutting methods find the invisible seams by dynamic programming or assign composition labels by graph-cut optimization. The masks used for fusion in these methods only contain values of 0 or 1.

However, for a learning system, the predicted masks with strict integers would prevent gradients from back-propagation. Moreover, the masks with strict integers could easily produce discontinuous contents in the composited results. Therefore, we define the values of the masks to be float and propose a smoothness constraint on the stitched image (Eq. \textcolor{red}{12} of the manuscript) to encourage the smooth transition on both sides of this ``seam".
Fig. \ref{comp_diff} shows the masks from seam cutting \cite{li2018perception} and ours, where our ``seam" is significantly wider. That is why we cannot quantitatively evaluate our composition in traditional metrics.


\begin{figure}[t]
   \begin{center}
      \subfloat[Robustness analysis of warp.]{\includegraphics[width=0.43\textwidth]{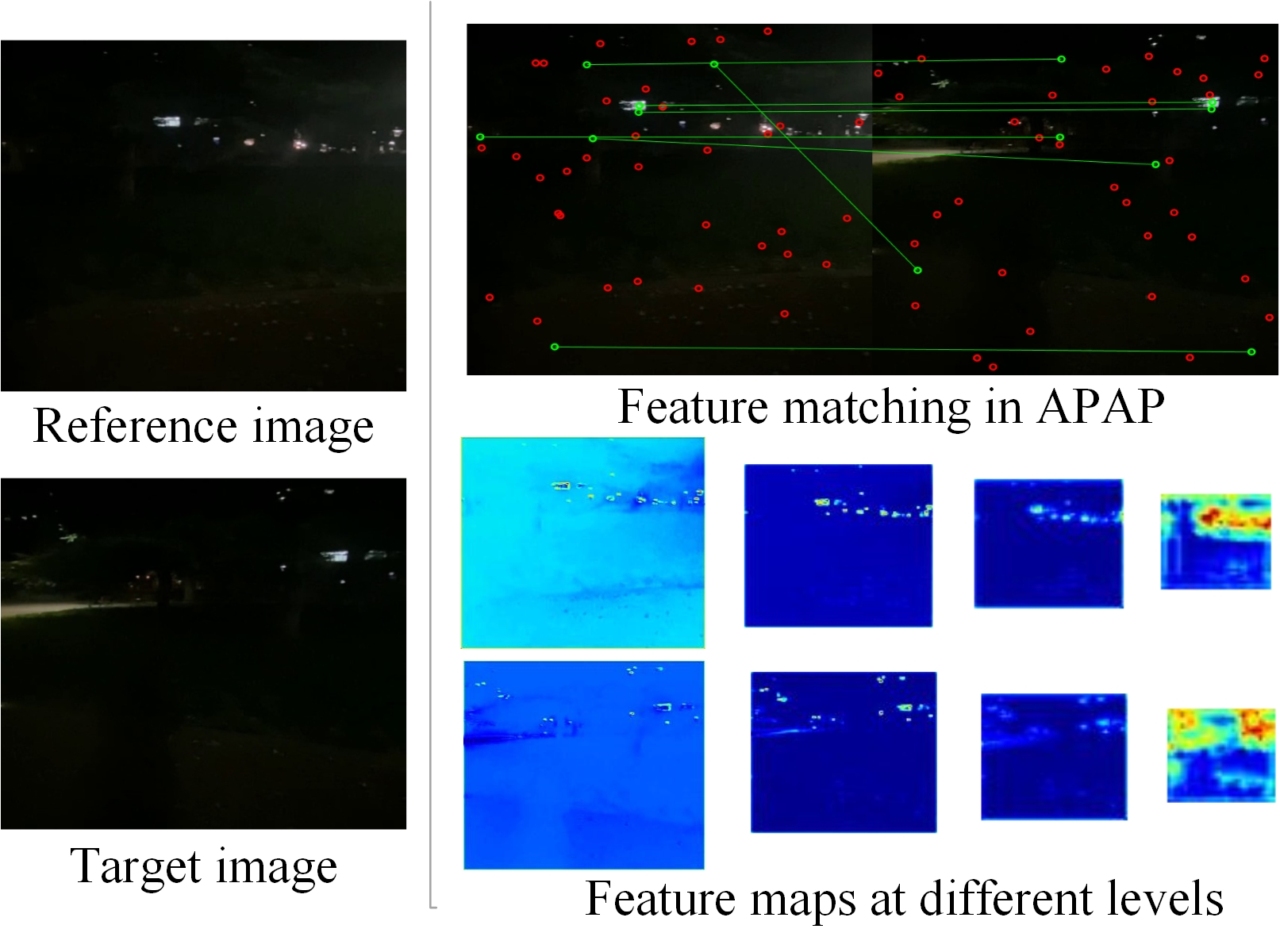}}
      \vspace{-0.3cm}
      \quad
      \subfloat[Robustness analysis of composition.]{\includegraphics[width=0.43\textwidth]{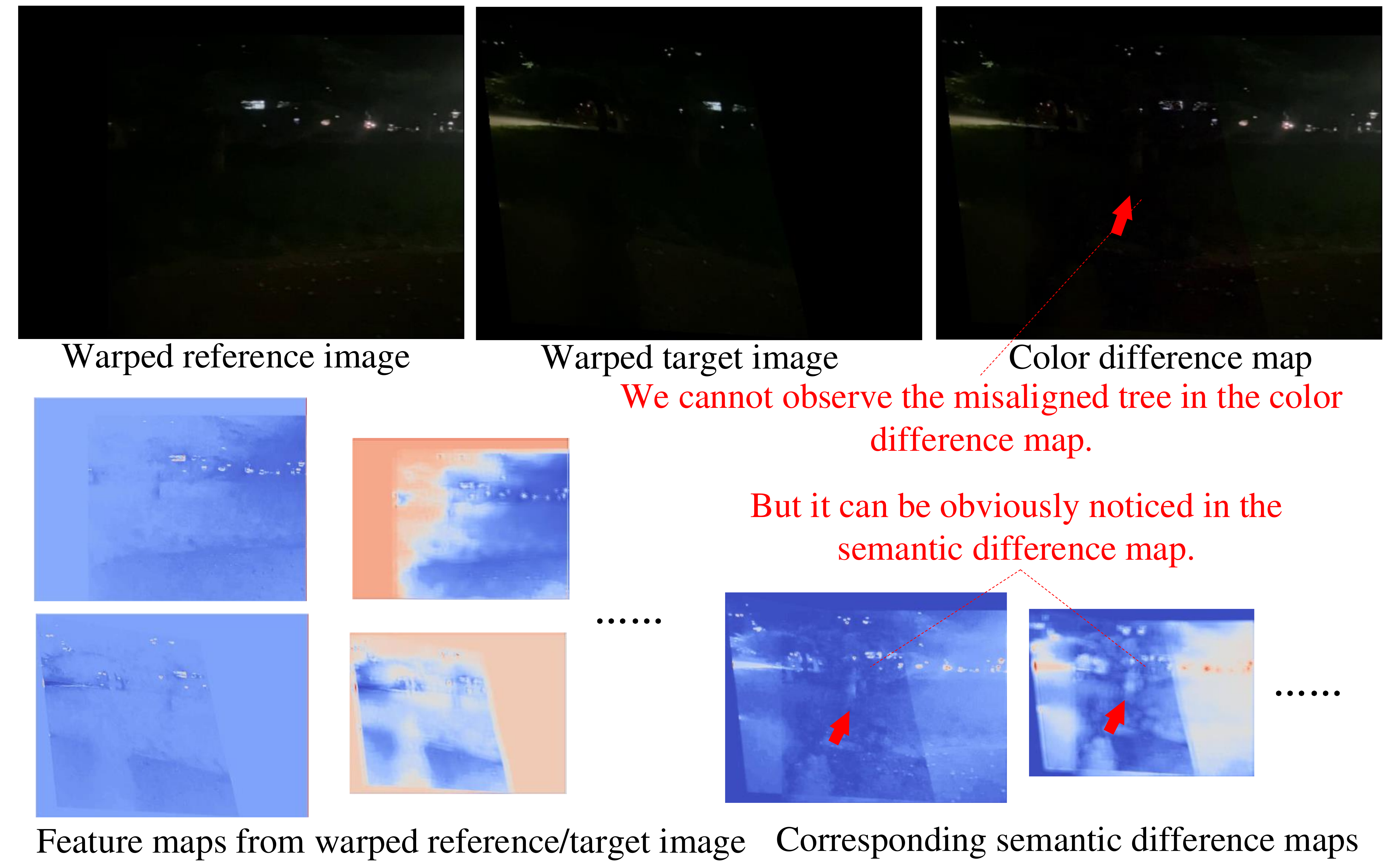}}
     \vspace{-0.5cm}
   \end{center}
   \caption{Robustness analysis.}
   \vspace{-0.5cm}
\label{fig2}
\end{figure}


\section{Analysis}
\label{analysis}
\subsection{Analysis on Robustness}
\label{sec:31}
\textbf{Warp:}
We argue that the proposed method is more robust than traditional solutions, especially in challenging cases. To illustrate this, we compare our method with APAP \cite{zaragoza2013projective}, which represents traditional solutions. In Fig. \ref{fig2}\textcolor{red}{a}, we show a challenging case with extremely low light. APAP extracts SIFT keypoints, which are marked using red or green circles. RANSAC is then used to remove the outliers (red circles), and the green line indicates matched keypoints. As shown in Fig. \ref{fig2}\textcolor{red}{a}, the keypoints are very sparse, and some keypoints are even mismatched, which can easily lead to stitching failure. In contrast, our solution extracts semantic feature maps, which become increasingly evident with the increase of network layers, contributing to our robustness.

\textbf{Composition:}
Regarding composition, existing seam-cutting methods mainly rely on color difference or other pixel-level energy maps. However, these maps often lose some essential content in challenging cases, such as low light. Fig. \ref{fig2}\textcolor{red}{b} displays an example where the tree (highlighted by red arrows) is missing in the color difference map. The proposed deep composition method overcomes this issue by extracting semantic difference maps, even though it is trained with color difference. Through training with extensive samples (both simple cases and challenging cases), the composition network is capable of perceiving the semantic difference even in low-light scenes. We illustrate the extracted feature maps and semantic residuals of the composition network in Fig. \ref{fig2}\textcolor{red}{b}, where the tree can be obviously noticed in semantic difference maps.

\begin{figure}[!t]
  \centering
  \includegraphics[width=0.45\textwidth, height=0.25\textheight]{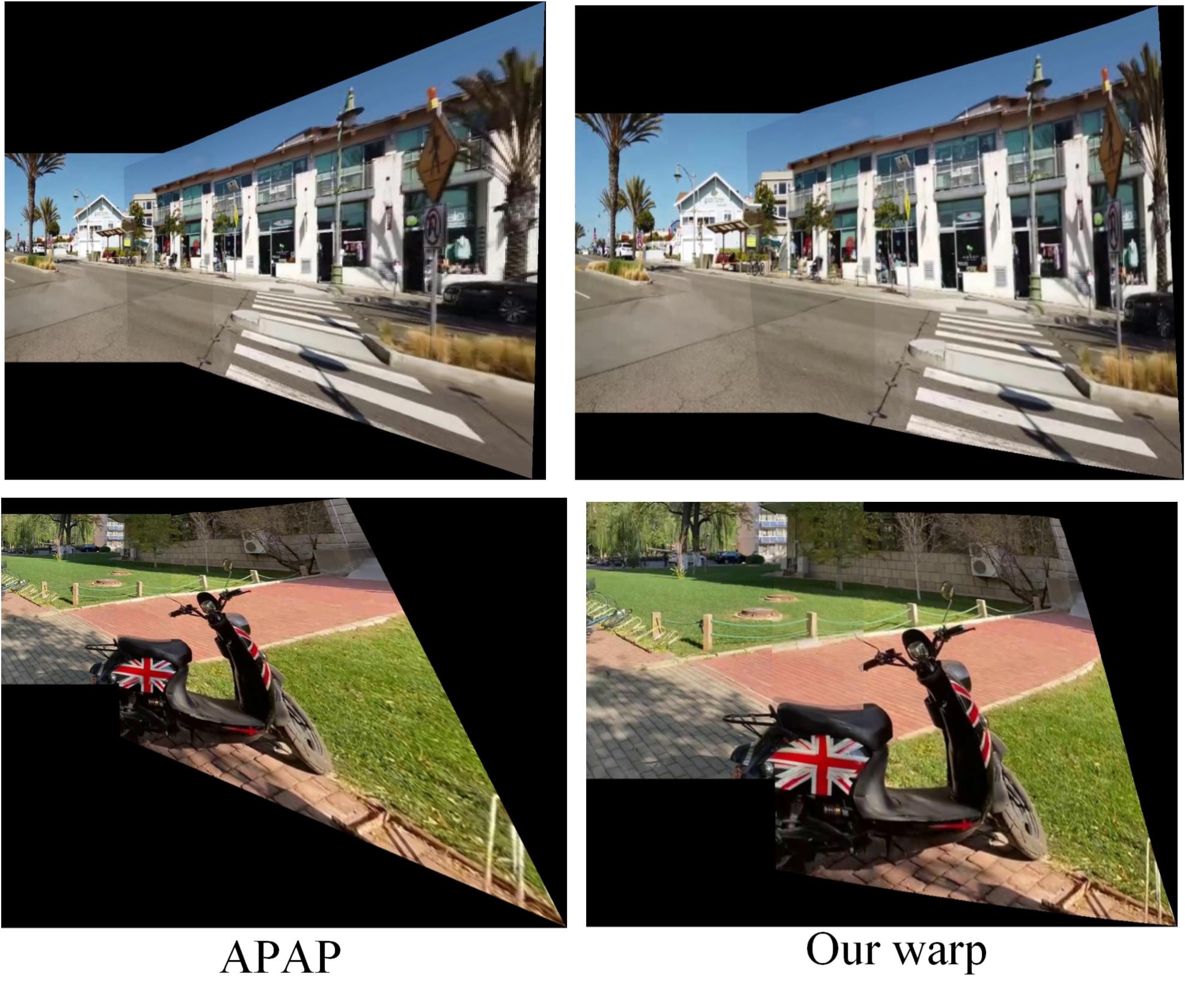}
  \vspace{-0.3cm}
  \caption{Projective distortion: APAP vs. ours. These instances are from UDIS-D dataset\cite{nie2021unsupervised}.}
  \label{fig3}
  \vspace{-0.3cm}
\end{figure}

\subsection{Analysis on Projective Distortion}
\label{sec:32}
Compared with other warps, our warp produces fewer projective distortions. We analyze the phenomenon from two perspectives:

i) Traditional methods estimate the warp from matched features. However, these features are usually distributed in some texture-rich local areas, so that the warp aligns well with these regions and overlooks other overlapping areas. Compared with them, our objective goal is to align all the pixels in overlapping regions (Eq. \textcolor{red}{6} of the manuscript). Therefore, our warp produces less projective distortions.

ii) To further eliminate projective distortions, we design an intra-grid constraint (Eq. \textcolor{red}{7} of the manuscript) to prevent the deformed mesh from scaling dramatically.

\begin{figure}[!t]
  \centering
  \includegraphics[width=0.35\textwidth]{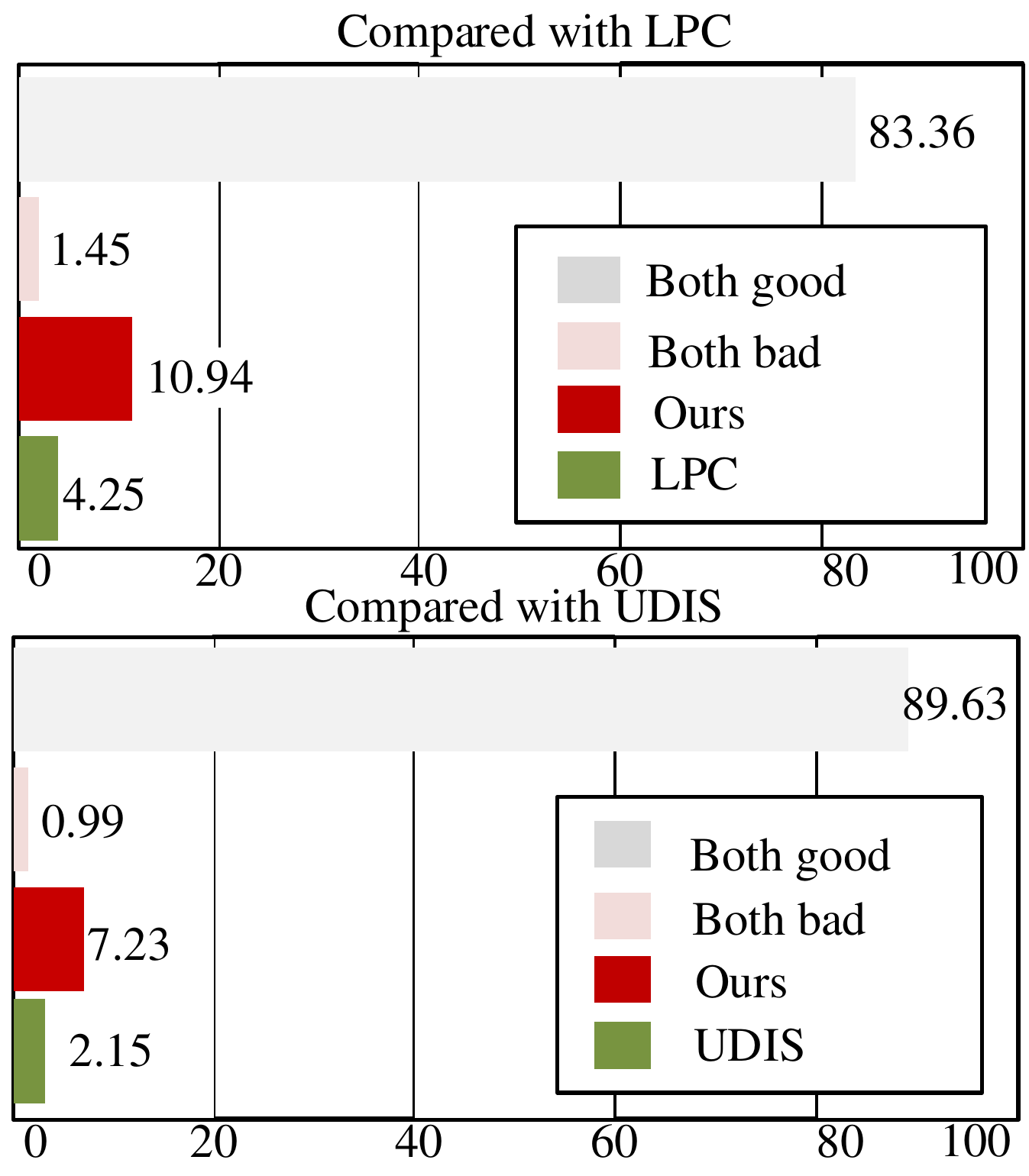}
  \vspace{-0.3cm}
  \caption{User study of visual preferences with existing SoTA solutions. The results are presented in percentage and averaged on 20 participants. }
  \label{user}
  \vspace{-0.4cm}
\end{figure}

\section{More Results}
\label{results}

\begin{figure}[!t]
  \centering
  \includegraphics[width=0.35\textwidth]{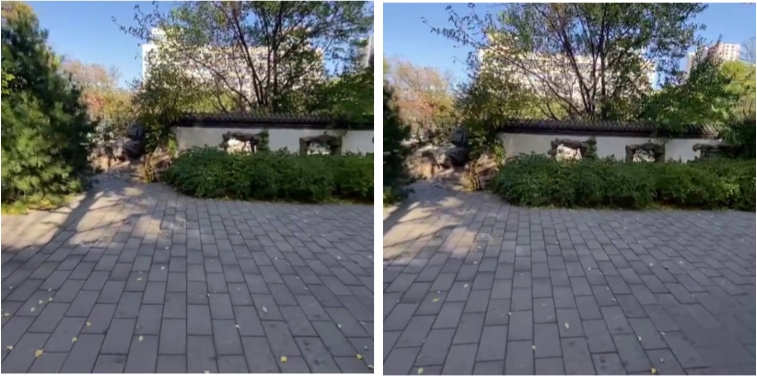}
  \vspace{-0.3cm}
  \caption{The input images of Fig. \textcolor{red}{4} in the manuscript.}
  \label{fig4}
  \vspace{-0.3cm}
\end{figure}

\begin{figure}[!t]
  \centering
  \includegraphics[width=0.35\textwidth, height=7cm]{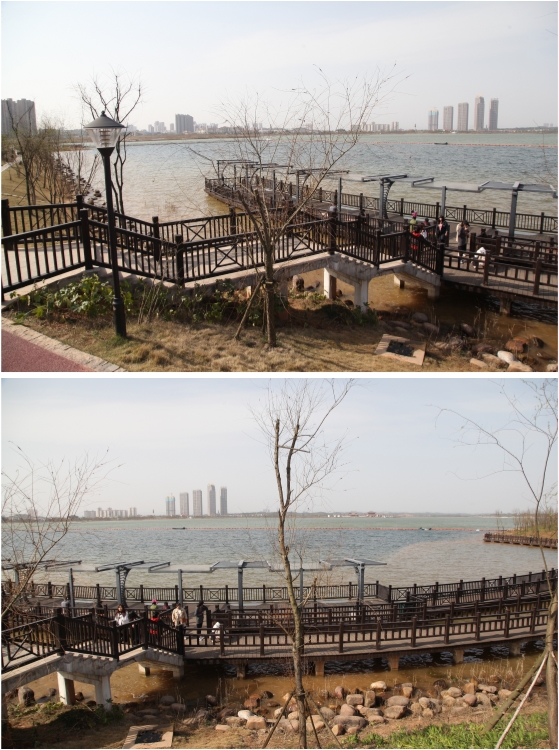}
  \vspace{-0.3cm}
  \caption{The input images of Fig. \textcolor{red}{5} in the manuscript.}
  \label{fig5}
  \vspace{-0.3cm}
\end{figure}

\subsection{Results of Warp}
\label{sec:41}
The Fig. \ref{fig4},\ref{fig5} of this material are the inputs of Fig. \textcolor{red}{4}, \textcolor{red}{5} in the manuscript. We demonstrate more results of warp on UDIS-D dataset and other datasets in Fig. \ref{fig6} and Fig. \ref{fig7}.

\subsection{Results of Composition}
\label{sec:42}
Here, we illustrate more comparative results of large-parallax composition in Fig. \ref{fig8}. To highlight the parallax artifacts intuitively, we use SIFT+RANSAC to align input images and blend the results with average fusion for reference. Then we compare our results with SoTA composition methods (perception-based seam cutting \cite{li2018perception} and reconstruction \cite{nie2021unsupervised}) in UDIS-D \cite{nie2021unsupervised} and other large-parallax datasets \cite{lin2016seagull}.

\subsection{Results of Complete Solutions}
\label{sec:44}
Then, we compare our complete framework with other SoTA solutions (LPC \cite{jia2021leveraging} and UDIS \cite{nie2021unsupervised}) with seam cutting or reconstruction as their post-processing operations. The qualitative results are shown in Fig. \ref{fig9}.

Moreover, we strictly follow the experimental setup in UDIS and conduct user studies to test visual preferences. The participants include 10 volunteers with computer vision backgrounds and 10 outside this community.
Specifically, we compare our method with LPC \cite{jia2021leveraging} and UDIS \cite{nie2021unsupervised} one by one. At each time, four images are shown on one screen: the inputs, our stitched result, and the result from LPC/UDIS. The results of ours and the other method are illustrated in random order each time. The user is allowed to zoom in on the images and is required to answer which result is preferred. In the case of “no preference,” the user needs to answer whether the two results are “both good” or “both bad”. The studies are carried out in the testing set of UDIS-D \cite{nie2021unsupervised}, which means every user has to compare each method with ours in 1,106 images.
The results are shown in Fig. \ref{user}.

Besides, we demonstrate more results in traditional datasets \cite{li2017parallax, lin2016seagull, zaragoza2013projective,  gao2011constructing, zhang2014parallax} in Fig. \ref{fig-cross}. Our solution can generate natural and seamless results in different scenes with various resolutions and parallax.
Also, we promise to release all subjective results, including 1,106 images in UDIS-D and others in traditional datasets.

\subsection{Results of Challenging Scenes}
We also demonstrate more results in some challenging scenes, such as low texture, low light, etc. As shown in Fig. \ref{challenge}, the traditional scheme fails to stitch these images due to the lack of geometric features. In contrast, our solution succeeds (the reason is discussed in Section \ref{sec:31}).

\begin{table}[!t]
  \centering
  \caption{Ablation studies of alignment performance on UDIS-D dataset\cite{nie2021unsupervised}. With the distortion term ($\ell_{inter}$+$\ell_{intra}$), the alignment performance decrease little.}
  \vspace{-0.2cm}
  \scalebox{0.95}{
  \begin{tabular}{llll}
   \toprule
   & \makecell[c]{Loss} &\makecell[c]{PSNR}&\makecell[c]{SSIM} \\
   \cline{1-4}
 1 & \makecell[c]{w/o $\ell_{inter}$+$\ell_{intra}$} &\makecell[c]{25.54}&\makecell[c]{0.841} \\
 2 & \makecell[c]{w/o $\ell_{intra}$} &\makecell[c]{25.53}&\makecell[c]{0.840} \\
 3 & \makecell[c]{w/o $\ell_{inter}$} &\makecell[c]{25.48}&\makecell[c]{0.839} \\
 4 & \makecell[c]{Our warp} &\makecell[c]{25.43}&\makecell[c]{0.838} \\

      \bottomrule
   \end{tabular}
  }
   \vspace{-0.1cm}
   \label{tab11}
   \end{table}

\begin{table}[!t]
  \centering
  \caption{The superiority of combining TPS with homography. The experiments are conducted on UDIS-D dataset.}
  \vspace{-0.2cm}
  \scalebox{0.95}{
  \begin{tabular}{llll}
   \toprule
   & \makecell[c]{Architecture} &\makecell[c]{PSNR}&\makecell[c]{SSIM} \\
   \cline{1-4}
 1 & \makecell[c]{Homography + Homography} &\makecell[c]{24.46}&\makecell[c]{0.802} \\
 2 & \makecell[c]{TPS + TPS} &\makecell[c]{25.31}&\makecell[c]{0.836} \\
 3 & \makecell[c]{Homography + TPS} &\makecell[c]{25.43}&\makecell[c]{0.838} \\
      \bottomrule
   \end{tabular}
  }
   \vspace{-0.1cm}
   \label{tab22}
   \end{table}

\subsection{Ablation Studies}
\label{sec:43}
As shown in Fig. \textcolor{red}{7} of the manuscript, the distortion constraints preserve the shape effectively. Also, it produces little negative impact on alignment. The quantitative results are shown in Table \ref{tab11}, where the SSIM merely decreases 0.03 when we adopt these shape-preserving constraints.

Besides, we demonstrate the superiority of combining TPS with homography in Table \ref{tab22}. Compared with only homography, the combination can significantly improve the alignment performance. Compared with only TPS, the combination reaches slightly better performance with less computational cost.


\begin{figure}[!t]
  \centering
  \includegraphics[width=0.48\textwidth]{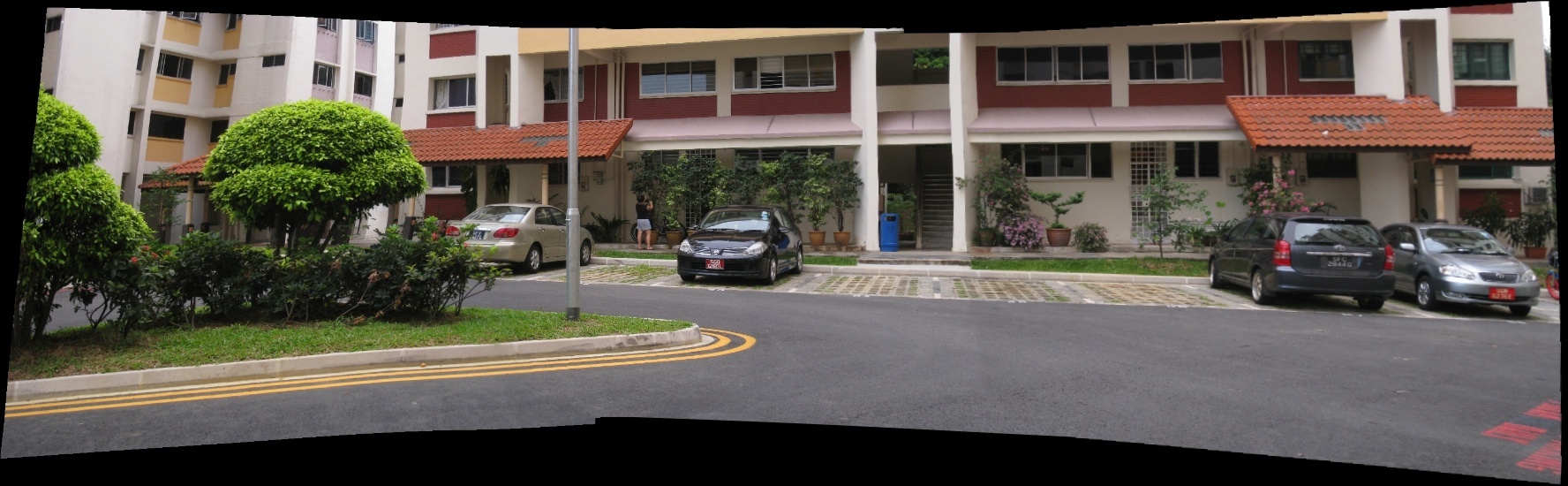}
  \vspace{-0.3cm}
  \caption{Stitching four images from the traditional dataset\cite{gao2013seam}.}
  \label{mis}
  \vspace{-0.3cm}
\end{figure}

\subsection{Multi-Image Stitching}
Most stitching methods (e.g., LPC\cite{jia2021leveraging}, UDIS\cite{nie2021unsupervised}) focus on stitching two images, and so do ours. However, stitching multiple images can be generalized by performing multiple pairwise stitching. Here, we show a case of stitching 4 images in Fig. \ref{mis}.

\begin{figure}[t]
   \begin{center}
      \subfloat[Low-texture cases.]{\includegraphics[width=0.43\textwidth, height=0.2\textheight]{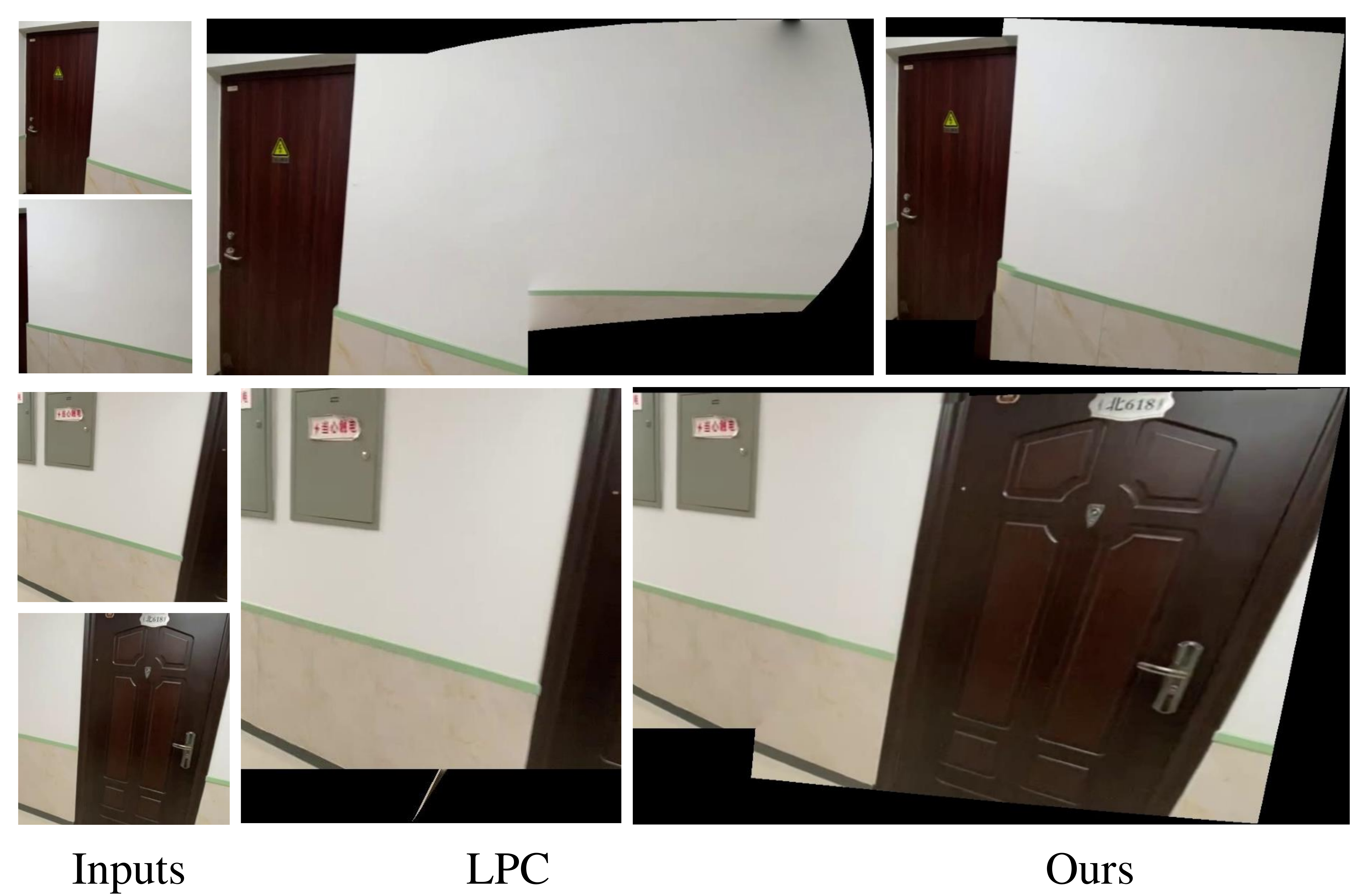}}
      \vspace{-0.3cm}
      \quad
      \subfloat[A case in the dark. Top: the original images (inputs and results). Bottom: images after enhancement for better observation.]{\includegraphics[width=0.43\textwidth, height=0.2\textheight]{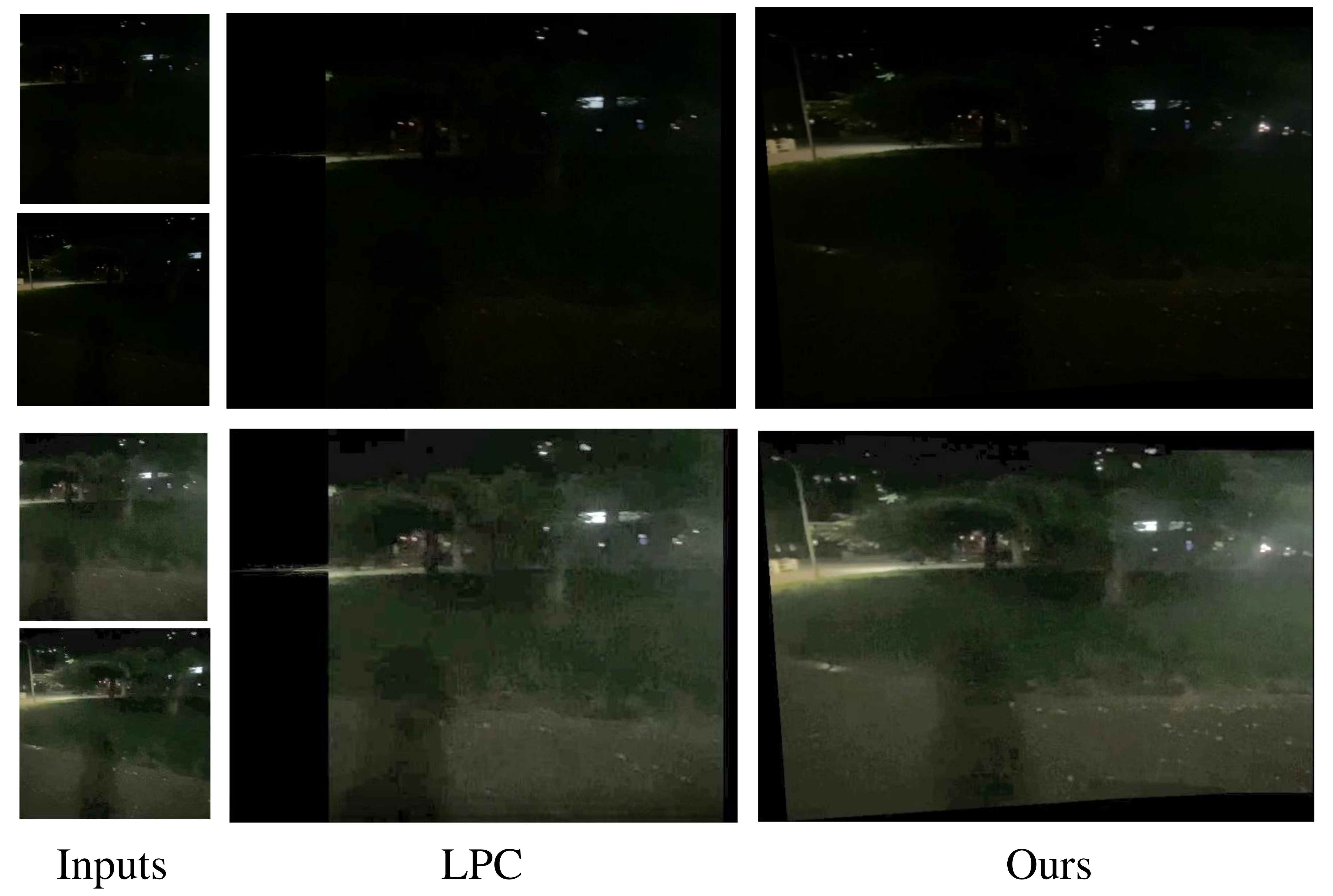}}
     \vspace{-0.5cm}
   \end{center}
   \caption{Results of challenging scenes. Traditional methods fail in these scenes due to the lack of geometric features. All the cases are from UDIS-D dataset \cite{nie2021unsupervised}}
   \vspace{-0.5cm}
\label{challenge}
\end{figure}



\begin{figure*}[!t]
  \centering
  \includegraphics[width=0.9\textwidth]{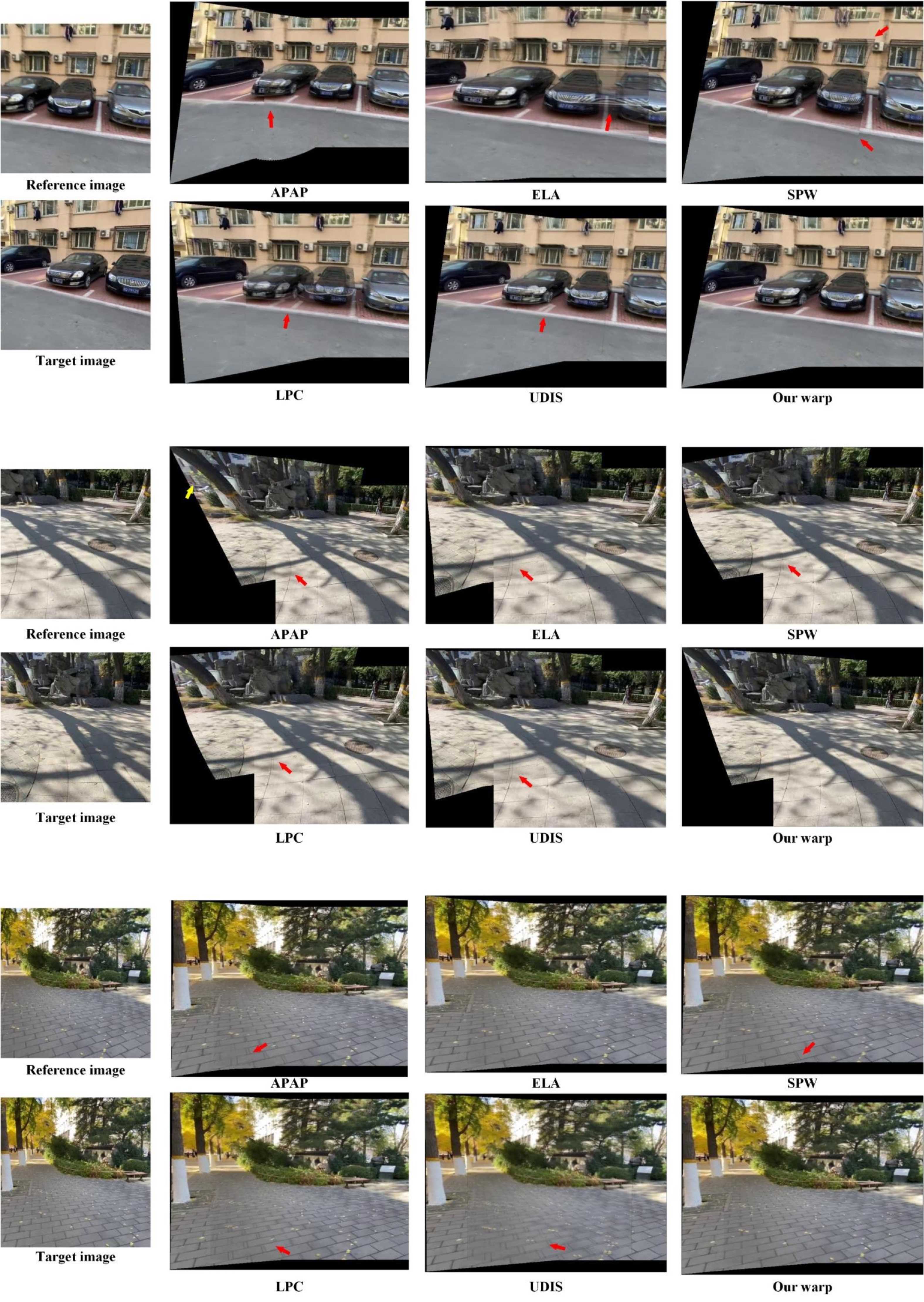}
  \vspace{-0.3cm}
  \caption{Comparative results of warp on UDIS-D dataset\cite{nie2021unsupervised}. The red arrows highlight the artifacts.}
  \label{fig6}
  \vspace{-0.3cm}
\end{figure*}

\begin{figure*}[!t]
  \centering
  \includegraphics[width=0.9\textwidth]{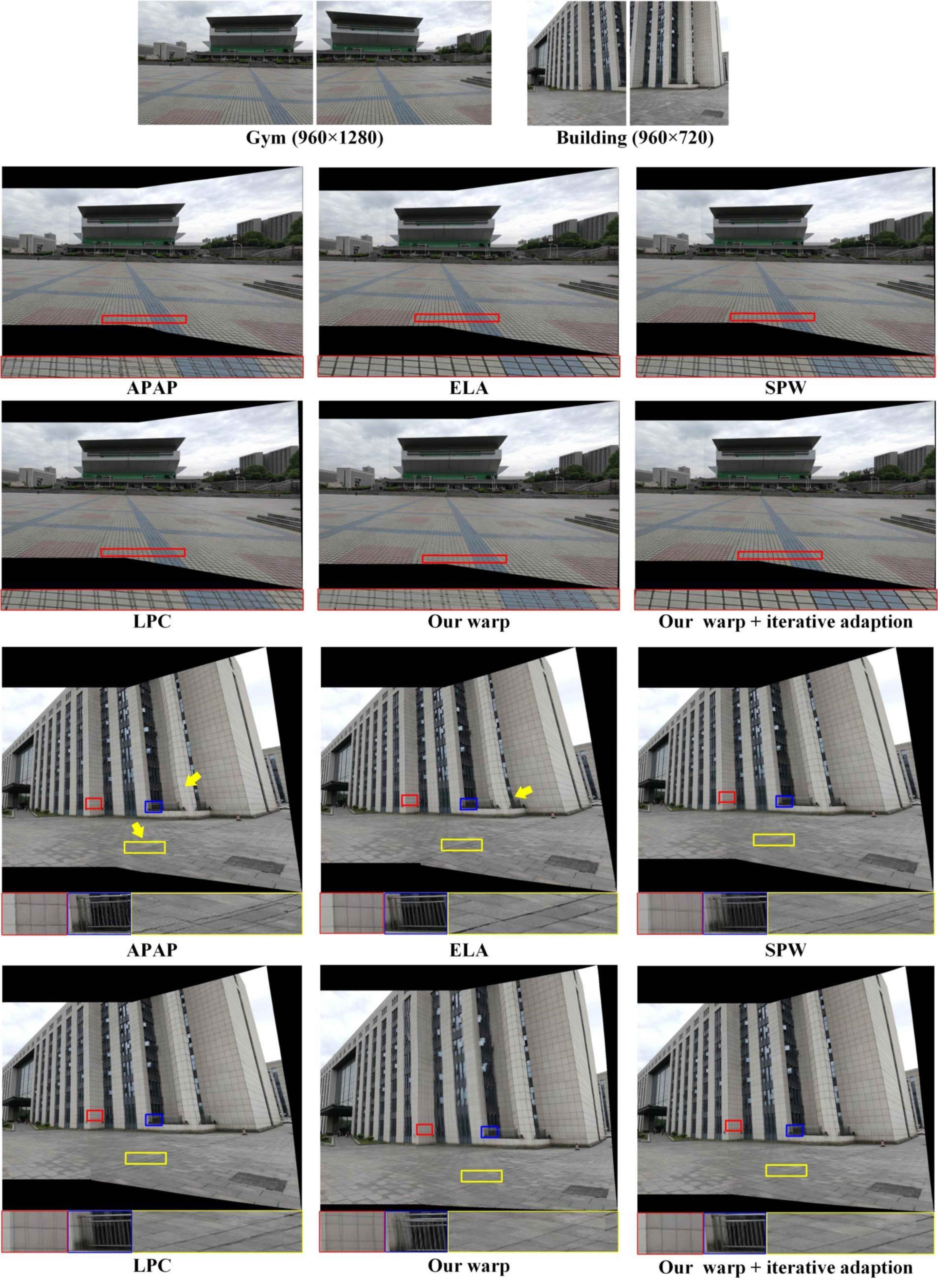}
  \vspace{-0.3cm}
  \caption{Comparative results of warp in cross-dataset cases\cite{li2017parallax}.The arrows highlight the distortions.}
  \label{fig7}
  \vspace{-0.3cm}
\end{figure*}



\begin{figure*}[t]
   \begin{center}
      \subfloat[Comparison of composition in UDIS-D dataset\cite{nie2021unsupervised}.]{\includegraphics[width=0.95\textwidth]{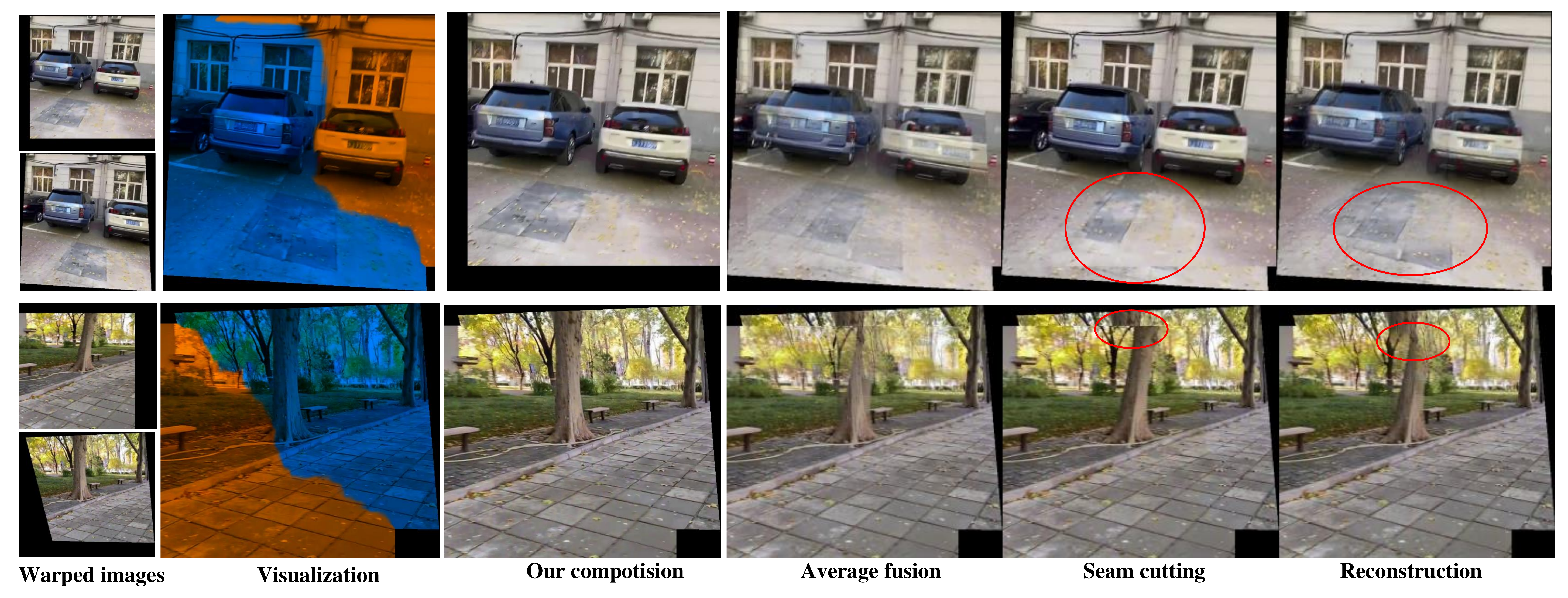}}
      \vspace{-0.3cm}
      \quad
      \subfloat[Comparison of composition in traditional large-parallax dataset\cite{lin2016seagull}.]{\includegraphics[width=0.95\textwidth]{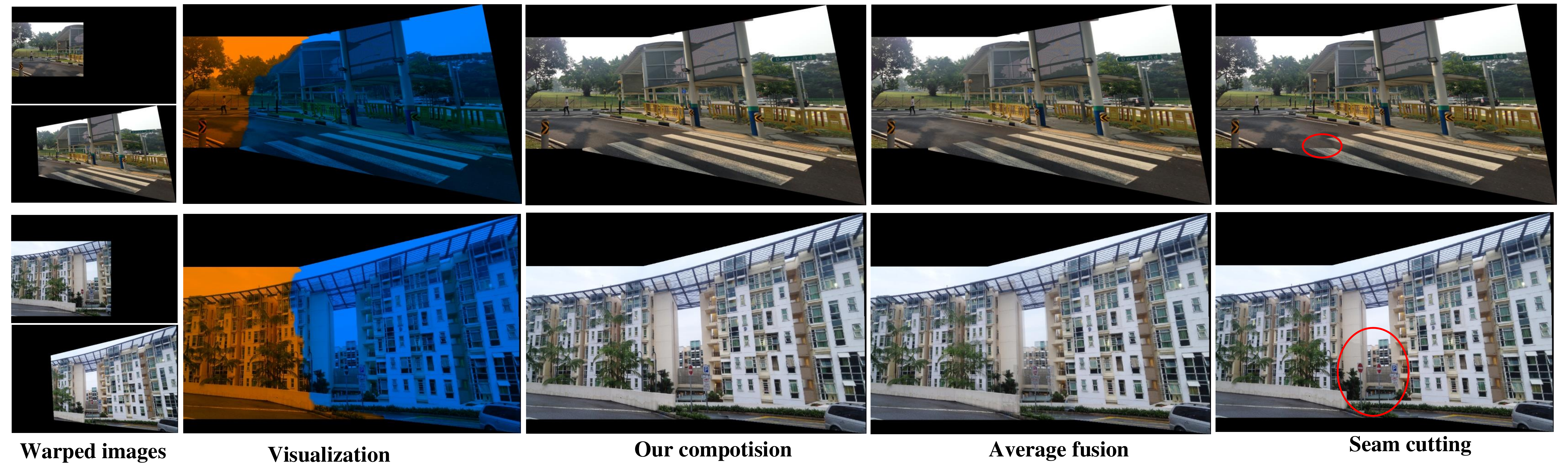}}
     \vspace{-0.5cm}
   \end{center}
   \caption{Comparative results of composition. We warp large-parallax cases using SIFT+RANSAC and all the composition methods take the warped images as input. The red circles highlight the seam discontinuity or blur.}
   \vspace{-0.1cm}
\label{fig8}
\end{figure*}

\begin{figure*}[!t]
  \centering
  \includegraphics[width=0.95\textwidth]{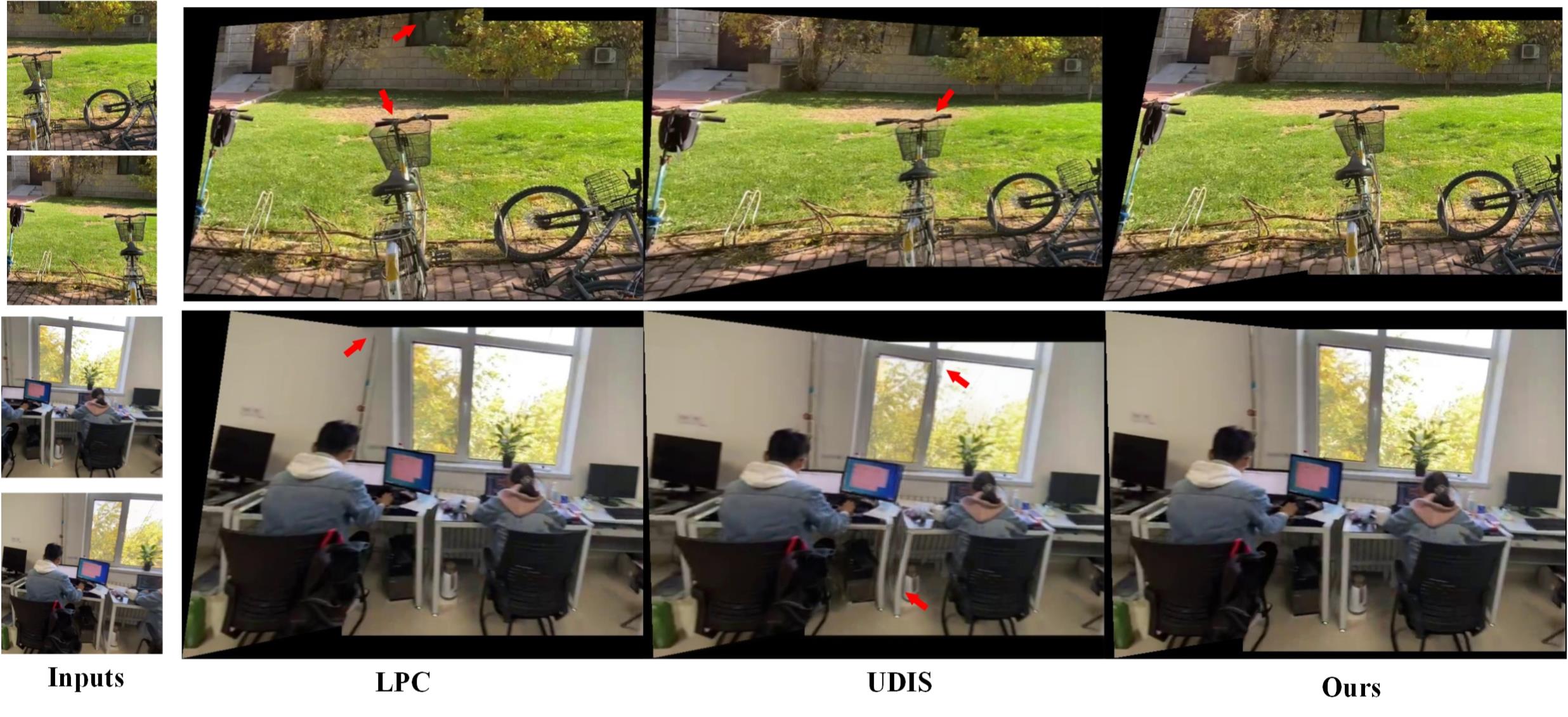}
  \vspace{-0.3cm}
  \caption{Comparative results of complete stitching frameworks. LPC\cite{jia2021leveraging} and UDIS\cite{nie2021unsupervised} leverage perception-based seam cutting\cite{li2018perception} and reconstruction\cite{nie2021unsupervised}as the composition methods.}
  \label{fig9}
  \vspace{-0.3cm}
\end{figure*}

\begin{figure*}[!t]
  \centering
  \includegraphics[width=0.94\textwidth, height =0.97\textheight]{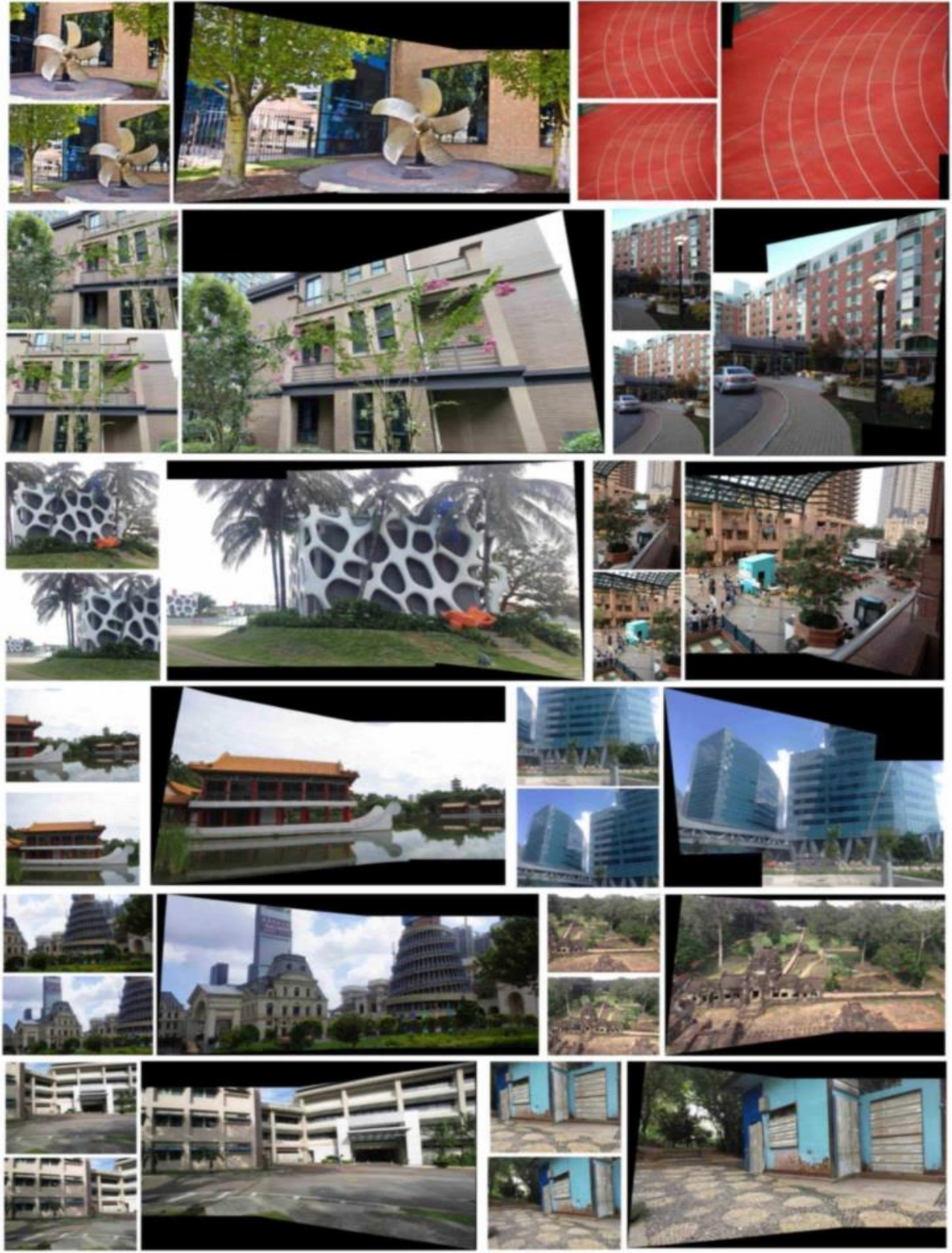}
  \vspace{-0.3cm}
  \caption{More results on traditional datasets \cite{li2017parallax, lin2016seagull, zaragoza2013projective,  gao2011constructing, zhang2014parallax}. The proposed method demonstrates good generalization in other scenes with various occlusions and parallax. }
  \label{fig-cross}
  \vspace{-0.3cm}
\end{figure*}

\begin{figure*}[!t]
  \centering
  \includegraphics[width=0.97\textwidth]{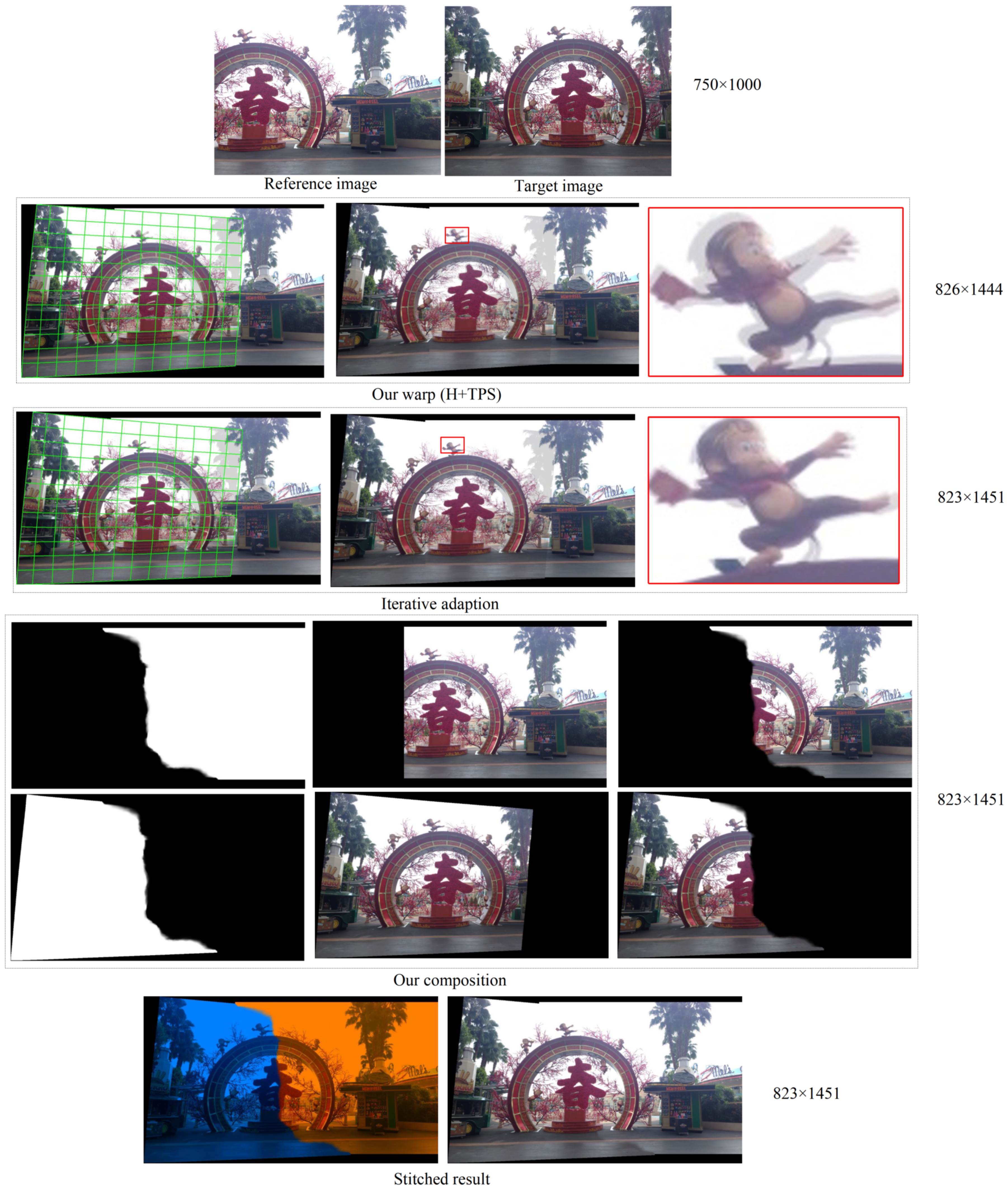}
  \vspace{-0.3cm}
  \caption{The complete pipeline of the proposed stitching framework. We show our intermediate result in a large-parallax cross-dataset case\cite{lin2016seagull}. We link the predicted control points to form a mesh for clear visualization. Note that for the images from UDIS-D dataset\cite{nie2021unsupervised}, we do not conduct warp adaption iterations.}
  \label{fig10}
  \vspace{-0.3cm}
\end{figure*}


\end{document}